\documentclass[10pt,journal,compsoc]{IEEEtran}
\pdfoutput=1
\usepackage{booktabs}
\usepackage{amsfonts}
\usepackage{graphicx}
\usepackage{multirow}
\usepackage{bookmark}
\usepackage{amsmath}
\usepackage{enumitem}

\usepackage[usenames]{color}
\usepackage{algorithm}
\usepackage{algpseudocode}
\usepackage{float}
\usepackage{amsthm}
\newtheorem{theorem}{Theorem}
\newtheorem{corollary}{Corollary}
\usepackage{cases}
\usepackage{subcaption}
\usepackage{wrapfig}
\usepackage{bm}
\usepackage{threeparttable}

\ifCLASSOPTIONcompsoc
  \usepackage[nocompress]{cite}
\else
  \usepackage{cite}
\fi
\begin{document}
\title{Multi-Head Encoding for Extreme Label Classification}

\author{Daojun Liang,~\IEEEmembership{Graduate~Student~Member,~IEEE,} 
        Haixia Zhang,~\IEEEmembership{Senior~Member,~IEEE,} \\
        Dongfeng Yuan,~\IEEEmembership{Senior~Member,~IEEE,} \ and
        Minggao Zhang% <-this % stops a space
\IEEEcompsocitemizethanks{\IEEEcompsocthanksitem D. Liang, H. Zhang, D. Yuan and M. Zhang are all with Shandong Provincial Key Laboratory of Wireless Communication Technologies, Shandong University, China. (e-mail: liangdaojun@mail.sdu.edu.cn; haixia.zhang@sdu.edu.cn; dfyuan@sdu.edu.cn; mgzhang@sdu.edu.cn) \protect
\IEEEcompsocthanksitem D. Liang is also with School of Information Science and Engineering, Shandong University, Qingdao, 266237, China.
\IEEEcompsocthanksitem H. Zhang and M. Zhang are also with School of Control Science and Engineering, Shandong University,  Jinan, 250061, China.
\IEEEcompsocthanksitem D. Yuan is also with School of Qilu Transportation, Shandong University, Jinan, 250002, China.
}% <-this % stops an unwanted space
%\thanks{Manuscript received xxx, xxx; revised xxx, xxx.}
\thanks{Corresponding author: Haixia Zhang}
% \thanks{ This work was supported in part by the Project of International Cooperation and Exchanges NSFC under Grant No. 61860206005, the National Natural Science Foundation of China under Grant No. 62171262 and No. 62271288, and the Key Research and Development (Major Scientific and Technological Innovation) Project of Shandong Province under Grant No. 2020CXGC010108.}
 }

\IEEEtitleabstractindextext{
\begin{abstract}
The number of categories of instances in the real world is normally huge, and each instance may contain multiple labels. To distinguish these massive labels utilizing machine learning, eXtreme Label Classification (XLC) has been established. 
However, as the number of categories increases, the number of parameters and nonlinear operations in the classifier also rises. This results in a Classifier Computational Overload Problem (CCOP).
To address this, we propose a Multi-Head Encoding (MHE) mechanism, which replaces the vanilla classifier with a multi-head classifier. 
During the training process, MHE decomposes extreme labels into the product of multiple short local labels, with each head trained on these local labels.
During testing, the predicted labels can be directly calculated from the local predictions of each head. This reduces the computational load geometrically.
Then, according to the characteristics of different XLC tasks, e.g., single-label, multi-label, and model pretraining tasks, three MHE-based implementations, i.e., Multi-Head Product, Multi-Head Cascade, and Multi-Head Sampling, are proposed to more effectively cope with CCOP.
Moreover, we theoretically demonstrate that MHE can achieve performance approximately equivalent to that of the vanilla classifier by generalizing the low-rank approximation problem from Frobenius-norm to Cross-Entropy.
Experimental results show that the proposed methods achieve state-of-the-art performance while significantly streamlining the training and inference processes of XLC tasks.
The source code has been made public at https://github.com/Anoise/MHE.
\end{abstract}

\begin{IEEEkeywords}
  Multi-Head Encoding, Extreme Label Classification, eXtreme Multi-label Classification.
\end{IEEEkeywords}}

\maketitle

\IEEEdisplaynontitleabstractindextext

\IEEEpeerreviewmaketitle

\IEEEraisesectionheading{\section{Introduction}\label{sec:introduction}}

\IEEEPARstart{I}{n} the real world, there exist millions of biological species, a myriad of non-living objects, and an immense natural language vocabulary.
To distinguish the categories of these massive instances, eXtreme Label Classification (XLC) \cite{bengio2003quick, morin2005hierarchical} is required, leading to a dramatic increase in the number of parameters and nonlinear operations in the classifier. 
This phenomenon, known as the Classifier Computational Overload Problem (CCOP), makes it challenging for existing machine learning methods using One-Hot Encoding (OHE) or multi-label learning algorithms to be practical due to the intractable computational and storage demands.

Currently, the primary tasks in XLC include eXtreme Single-Label Classification (XSLC), eXtreme Multi-Label Classification (XMLC), and model pretraining.
For XSLC, sampling-based \cite{bengio2003quick, bengio2008adaptive, jean2014using} and softmax-based \cite{morin2005hierarchical, chen2015strategies, kim2016character} methods are employed to train neural language models, reducing the complexity in computing the output.
For XMLC, e.g., multi-label text classification, many researchers utilize one-versus-all \cite{yen2016pd, babbar2017dismec, jain2016extreme, liu2017deep}, Hierarchical Label Tree (HLT) \cite{prabhu2014fastxml, prabhu2018parabel, khandagale2020bonsai, wydmuch2018no, yu2022XRlinear}, Label Clustering (LC) \cite{tagami2017annexml, you2019attentionxml, chang2020taming, jiang2021lightxml} etc., label preprocessing techniques to decompose extreme labels into a small and tractable label space.
For model pretraining tasks, e.g., face recognition, pretrained models must be trained on datasets containing millions of faces. Consequently, the authors in \cite{zhang2018accelerated} and \cite{an2021partial} employ hash forest or random sampling methods to approximate the original OHE.

\begin{figure}
  \centering
  \centerline{\includegraphics[width=\columnwidth]{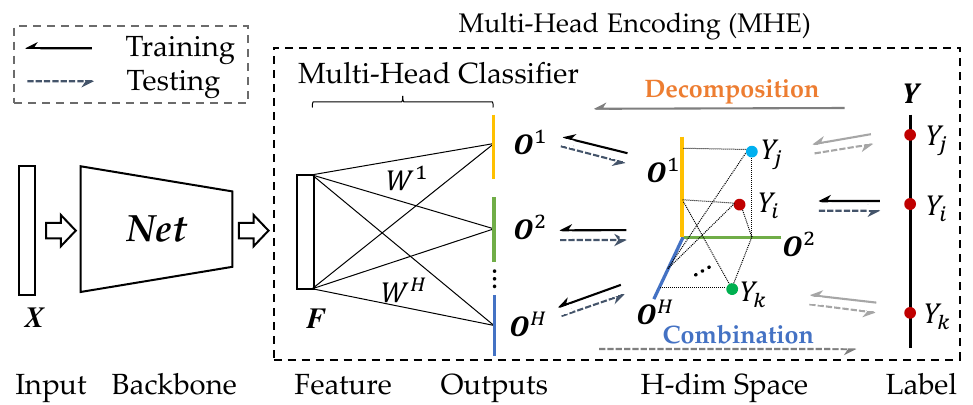}}
  \caption{The deep neural networks which are composed of three parts: input, backbone and classifier. In multi-head encoding, labels are decomposed onto the outputs of the multi-head classifier during training, and the outputs are combined in testing to obtain the predicted labels.}
  \label{fig0}
\end{figure}

Different from the above methods, as shown in Fig. \ref{fig0}, we decompose the original classifier into multiple heads and conceptualize extreme labels as points in a high-dimensional space.
During training, the coordinate components of an extreme label correspond to the local labels of each head. 
This process involves decomposing the extreme label into the product of multiple local labels, thereby geometrically reducing the encoding length of the extreme label.
During testing, each head contributes a coordinate component to form a point in the high-dimensional space, which can be projected onto the integer axis to obtain the extreme label. 
Since the extreme label can be calculated from the encoded information of the local labels, we term this mechanism Multi-Head Encoding (MHE).

% \begin{wrapfigure}{r}{0.5\columnwidth}
%   \begin{center}
%   \vspace{-15pt}
%     \includegraphics[width=0.5\columnwidth]{figs/workflow.pdf}
%   \end{center}
%   \vspace{-10pt}
%   \captionsetup{font={scriptsize},justification=justified}
%   \caption{The main workflow of the paper.}
%   \vspace{-10pt}
%   \label{fig0_workflow}
% \end{wrapfigure}

Based on their inference methods and application scenarios, MHE can be applied to various XLC tasks, such as XSLC, XMLC, and model pretraining.
We propose three algorithmic implementations of MHE.
First, a Multi-Head Product (MHP) algorithm is designed for XSLC. 
This algorithm directly employs the product operation to combine the classification heads, yielding rapid computation speed and commendable performance.
Second, a Multi-Head Cascade (MHC) algorithm is designed for XMLC. MHC also adopts the product operation, but a sequential cascade among heads is constructed to eliminate ambiguity in the multi-label representation.
Finally, a Multi-Head Sampling (MHS) algorithm is designed for model pretraining. MHS does not combine the multi-heads. Instead, only the local head corresponding to the ground truth label is trained each time.
The three algorithms have achieved considerable performance and speed advantages across various XLC tasks.

Then, to study the representation ability of MHE, we consider that the essence of MHE lies in its use as a low-rank approximation method, i.e., MHE approximates high-order extreme labels using multiple first-order heads.
Specifically, we generalize the low-rank approximation problem from the Frobenius-norm to the Cross-Entropy (CE) metric and demonstrate that CE enables the outputs of the multi-head classifier to closely approximate those of the vanilla classifier.
Two conclusions can be drawn from this theoretical analysis: 1) The performance gap between OHE and MHE is considerably small. 2) Label preprocessing techniques, e.g., HLT and label clustering, are not necessary since the low-rank approximation remains independent of label positioning.
These conclusions are also verified by our experiments.

\textbf{The main contributions are summarized as follows:}
\begin{itemize}%[leftmargin=*]
  \item An MHE mechanism is proposed to address the challenge of CCOP within XLC tasks. This mechanism geometrically reduces computational complexity and significantly simplifies both the training and inference processes.
  \item Three MHE-based algorithms are designed for various XLC tasks. The experimental results demonstrate that the three algorithms achieve state-of-the-art (SOTA) performance, providing strong benchmarks.
  \item The low-rank approximation problem is generalized from the Frobenius-norm to CE to theoretically analyze the representation ability of MHE. 
  It is proven that the performance gap between OHE and MHE is considerably small, and label preprocessing techniques are not necessary.
\end{itemize}

The remainder of this paper is organized as follows: 
Related works are introduced in Section \ref{sec_rework}.
In Section \ref{sec_mhe}, we first introduce the definition of CCOP, and then delve into the MHE mechanism.
In Section \ref{sec_mhe_imp}, we present three algorithmic implementations of MHE for XSLC, XMLC, and model pretraining. Then, we offer a strategy to determine the number and length of the heads.
In Section \ref{sec_theorem_dis}, we generalize the low-rank approximation problem from Frobenius-norm to CE to theoretically analyze the representation ability of MHE.
Finally, experiments are presented in Section \ref{sec_exp}, followed by the discussions and conclusions in Sections \ref{sec_diss} and \ref{sec_con}.

\section{Related Work}
\label{sec_rework}

Researches on XLC can be summarized into four categories:

\textbf{Sampling-based Methods}. XLC was first adopted in the field of Natural Language Processing (NLP) to solve out-of-vocabulary problems. There are two ways to achieve this. One is to adopt importance sampling to calculate the softmax, and the other is first performing label encoding and then approximating the softmax. The sampling-based methods \cite{bengio2003quick, bengio2008adaptive, mikolov2013distributed, jean2014using} do not change the structure of the classifier but use the proposal distribution to sample negative instances to train the model. However, due to the large difference between the proposal and the real distributions, the number of instances in the training process increases rapidly, and its computational cost quickly exceeds that of the original model. For the face recognition task, Partial-FC proposed in \cite{an2021partial} uses a distributed sampling algorithm, which randomly samples a fixed proportion of instances samples to calculate the softmax probability. However, this sampling method is limited by the sampling rate, which greatly limits its applicability.

\textbf{Softmax-based Methods}. Research on softmax approximation in NLP includes Hierarchical Softmax (H-Softmax) \cite{morin2005hierarchical}, D-Softmax \cite{chen2015strategies}, CNN-Softmax \cite{kim2016character}, etc. These methods first rank the frequency of word occurrences and then encode the vocabulary through BytePair \cite{sennrich2015neural}, WordPiece \cite{kudo2018subword}, and SentencePiece \cite{kudo2018sentencepiece} to reduce computational complexity. In model pretraining, the authors in \cite{zhang2018accelerated} use a hash forest to approximate the selected activation categories for face recognition. This method recursively divides the label space into multiple subspaces and selects the maximum activations as the decision result. 
These methods either require statistical analysis of the vocabulary or only accelerate the training process of the model.

\textbf{One-Versus-All Methods}.
One-versus-all methods divide the extreme label space into multiple easy-to-handle subspaces and only process part of the label space during the training process. For example, some works \cite{yen2016pd, babbar2017dismec,jain2016extreme, liu2017deep} use one-versus-all methods to transform the XMLC problem into a binary classification problem. However, one-versus-all can only accelerate the training process of the model, and the complexity of the inference process is still linear in relation to the label space. There are also many studies using tree-based methods for label partitioning. For example, several works \cite{prabhu2014fastxml, prabhu2018parabel, khandagale2020bonsai, wydmuch2018no, yu2022XRlinear} use a logarithmic depth b-array HLT. However, they involve complex optimizations at node splitting, making it difficult to obtain cost-effective and scalable tree structures \cite{liu2021emerging}.

\textbf{Label Clustering Methods}.
For XMLC tasks, the cluster labels are first obtained via semantic embedding and a surrogate loss, and then the extreme labels are fine-classified within the clusters. Some works \cite{tagami2017annexml, you2019attentionxml, chang2020taming, jiang2021lightxml} use the probabilistic label tree for label clustering, an HLT-based bag-of-words method. Another work \cite{hang2021collaborative} introduces a graph autoencoder to encode the  semantic dependencies among labels into semantic label embeddings.
The key to these methods lies in label clustering, an essential yet exceedingly complex and time-consuming preprocessing step.

\begin{figure*}
  \centering
  \centerline{\includegraphics[width=\textwidth]{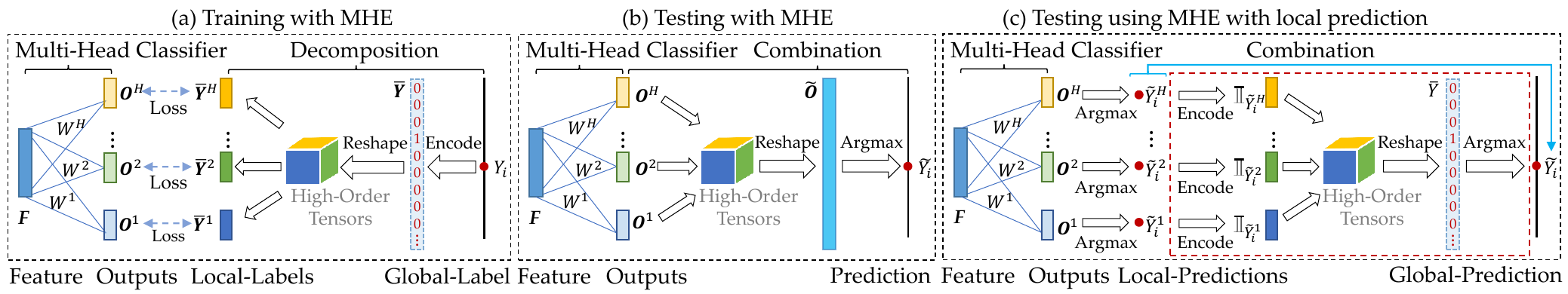}}
  \caption{The training and testing process of the multi-head classifier using MHE. (a) During training, the global label is decomposed into multiple local labels. (b) During testing, MHE works in the inverse manner of training. (c) The equivalent form of MHE for testing combines the local predictions of the multi-head classifier to obtain the global prediction. Note that the operations marked by the red dotted frame are used for ease of understanding, which are not required. } %(a) and (b) are equivalent from a low-rank approximation perspective.
  \label{fig1}
\end{figure*}

\section{Multi-Head Encoding (MHE)}
\label{sec_mhe}

\subsection{Notations}
For the sake of denotation simplicity, the variables are denoted using different fonts: 1) Constants are denoted by capital letters, e.g., $Y_i$ is a constant representing the category of the $i$-th sample, and $C$ represents the total number of categories. 
2) Vectors are denoted by bold capital letters, e.g., $\bm Y$ is the label set of $Y_i$, and $\bm{O}$ is the output of the head. 
3) Matrices are denoted by calligraphy capital letters, e.g., $\mathcal{W}$ represents the weight of the classifier, and $\mathcal{F}$ is the matrix of feature sets $\bm F$.
4) Tensors are also denoted by calligraphy capital letters with the superscript indicating their order, e.g., $\mathcal{Y}^{1,\cdots, H}$ represents an $H$-order tensor. %Through the above notations, readers can easily distinguish the shape of each variable. 

\subsection{Classifier Computational Overload Problem}

The deep neural network considered in this work is as shown in Fig. \ref{fig0}, which is composed of three parts: input, backbone, and classifier.
Assume that the given sample-label pairs are $\{\bm{X}_i, Y_i\}_{i=1}^N$, where $\bm{X} \in \mathbb{R}^{|\bm{X}|\times N}$ and $\bm{Y} \in \mathbb{R}^{C \times 1}$ are the sample set and label set, respectively. 
Let $\bar{\bm Y}_i$ denote the encoded (vectorized) label of $Y_i$. The backbone mainly includes multi-layer nonlinear neurons, denoted as $Net$. The feature output from $Net$ is denoted as $\bm F$, and the part that projects $\bm{F}$ into $\mathbb{R}^{C\times 1}$ through weight $\mathcal{W}$ is the classification head. The loss between output $\bm O$ and $\bar{\bm Y}$ is denoted as $\mathcal{L}$. Thus, the forward process of the network with single label can be formulated as
\begin{align}
    & \bm{F}_i =Net(\bm{X}_i), \label{eq1} \\
    & \bm{O}_i =\mathcal{W}\bm{F}_i+\bm{B},  \label{eq2} \\
    & \mathcal{L} = - \sum_i^N \bar{\bm{Y}}_i^T log(\sigma(\bm{O}_i)), \label{eq3}
\end{align}
where $\sigma$ is the softmax function, and $\bm{B} \in \mathbb{R}^{C\times 1}$ is the bias of the output layer. 
For XLC tasks, the length of $\bm{O}$ in Eq. \ref{eq2} must be equal to $C$, resulting in the Classifier Computational Overload Problem (CCOP) since $|\bm{O}|\gg |\bm{F}|$ \footnote[1]{$|\bm {F}|$ is generally on the order of 1K.} leads to computation being overweight.

\subsection{Label Decomposition}
\label{sec3_x}

Label decomposition in MHE involves decomposing the extreme label into multiple easy-to-handle local labels, which are then used to train neural networks.
To better understand this process, we can conceptualize the extreme label $Y_i$ as a point in high-dimensional space.
Then, its orthogonal coordinate components are used as local labels to train the neural network with multi-heads.
This process reduces the encoding length of the labels by reusing coordinate positions, thereby geometrically decreasing the computational load of the classifier.

The key to the label decomposition process is how to map $Y_i$ into an $H$-dimensional space. The solution proposed in this paper is to view $Y_i$ as a one-hot encoded vector $\mathbb{I}_{Y_i}$, as shown in Fig. \ref{fig1}a. Then it is reshaped into an $H$-dimensional tensor, $\bar{\mathcal{Y}}_i^{1,...,H}$. Note that since $\mathbb{I}_{Y_i}$ is a one-hot vector, $\bar{\mathcal{Y}}_i^{1,...,H}$ and its components, $\{\bar{\bm Y}_i^h\}_{h=1}^H$, are all one-hot encoded. Thus, the decomposition process of $Y_i$ can be formulated as
\begin{equation}
  \mathbb{I}_{Y_i} = \bar{\bm Y}_i^1 \otimes \bar{\bm Y}_i^2 \otimes \cdots \otimes \bar{\bm Y}_i^H, \label{eq4_}
\end{equation}
where $\otimes$ is the Kronecker product. Please refer to Appendix \ref{sec_ae_kron} for detailed examples. 
Equation \ref{eq4_} means that the extreme label is decomposed into the product of $H$ short one-hot vectors, each approximately of length $\sqrt[H]{C}$. Therefore, assigning each local label to each head to train the network will geometrically reduce the number of parameters in the classifier, thus solving the CCOP.

% Please refer to Section 4 for more details and Appendix 2 for the examples of label decomposition.

\begin{figure*}
  \centering
  \centerline{\includegraphics[width=\textwidth]{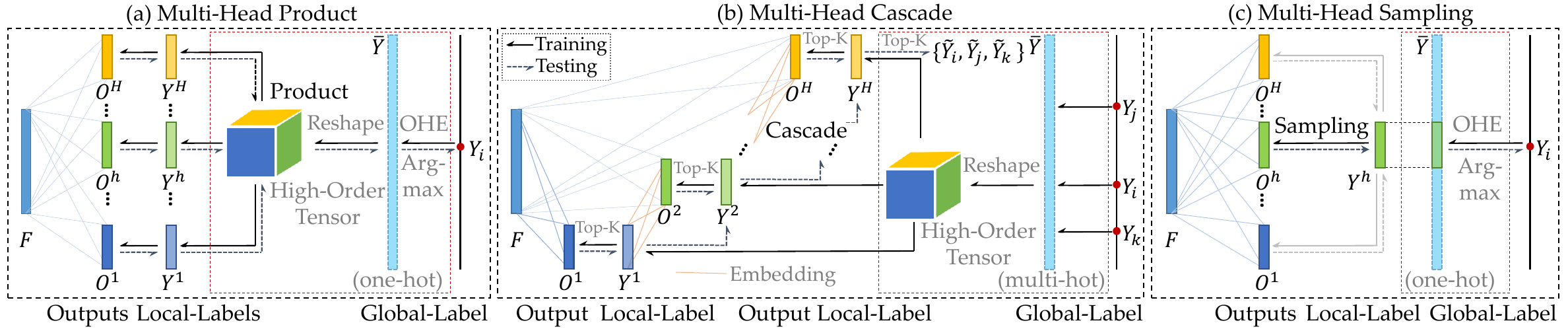}}
  \caption{Three MHE-based training and testing processes for XLC tasks. The part indicated by the red dotted frame is for ease of understanding, which is not required in practice.}
  \label{fig3}
\end{figure*}

\subsection{Multi-Head Combination}

The previous subsection showed how to decompose extreme labels into multiple short local labels to train the model. This subsection shows how to combine the outputs of the multi-heads to recover the global predicted label $\tilde{Y}_i$ (original extreme label) during testing.

In fact, the combination operation used during testing is the inverse of the decomposition operation used during training. 
As shown in Fig. \ref{fig1}b, if the output $\bm{O}_i^h$ of each head is viewed as a coordinate component, $\{\bm{O}_i^h\}_{h=1}^H$ can be producted to form the coordinates of a point in the $H$-dimensional space. Then, the predicted label $\tilde{Y}_i$ is obtained by projecting this point onto the integer axis as
\begin{align}
  \tilde{Y}_i = & \varLambda (\tilde{\bm O}_i) = \varLambda(\bm {O}_i^1 \otimes \bm {O}_i^2 \otimes \cdots \otimes \bm {O}_i^H), \label{eq5_}
\end{align}
where $\varLambda$ is the Argmax operation. 
Although $\tilde{Y}_i$ can be obtained by Eq. \ref{eq5_}, the overwhelming length of $\tilde{\bm O}_i$ ($|\tilde{\bm O}_i|=C$) and $H-1$ Kronecker product operations will consume huge computing and storage resources.
Therefore, it is necessary to simplify the inference process in Eq. \ref{eq5_}.

A desired solution is to leverage the local predicted labels to directly calculate the global predicted label,
\begin{align}
  \hat{Y}_i =  \varLambda(\bar{\bm Y}_i)  & = \varLambda( \mathbb{I}_{\tilde{Y}_i^1} \otimes \mathbb{I}_{\tilde{Y}_i^2} \otimes \cdots \otimes  \mathbb{I}_{\tilde{Y}_i^H}) \nonumber \\ 
  & = \varLambda( \mathbb{I}_{\varLambda(\bm{O}_i^1)} \otimes \mathbb{I}_{\varLambda(\bm{O}_i^2)} \otimes \cdots \otimes  \mathbb{I}_{\varLambda(\bm{O}_i^H)}) , \label{eq6_}
\end{align}
where $\mathbb{I}_{\varLambda(\cdot)}$ is OHE after a vector performed Argmax. 
It can be observed that $\tilde{Y}_i$ in Eq. \ref{eq6_} is the product of $H-1$ one-hot encoded vectors, which can be calculated from the local labels and their lengths as
\begin{align}
  \tilde{Y}_i = & f (\{\tilde{Y}_i^h, |O^h|\}_{h=1}^H) , \label{eq5_1}
\end{align} 
where $f$ is a function to be determined.

Although Eq. \ref{eq6_} can calculate the global predicted label by combining the local predicted labels, is it equivalent to Eq. \ref{eq5_}? i.e., is $\hat{Y}_i$ equal to $\tilde{Y}_i$?
Actually, the combination processes in Eq. \ref{eq6_} and Eq. \ref{eq5_} are equivalent, which can be proved by the following theorem.
\begin{theorem}
  For the outputs $\{\bm{O}^h\}_{h=1}^H$ of the multi-head classifier, we have \label{th1}
  \begin{align}
    \tilde{Y} & = \varLambda(\bm{O}^1 \otimes \bm{O}^2 \otimes \cdots \otimes \bm{O}^H)  \nonumber \\
    &  = \varLambda(\mathbb{I}_{\varLambda(\bm{O}^1)} \otimes \mathbb{I}_{\varLambda(\bm{O}^2)} \otimes \cdots \otimes  \mathbb{I}_{\varLambda(\bm{O}^H)}). \label{eq7_}
  \end{align}
\end{theorem}
The proof of Theorem \ref{th1} is given in Appendix \ref{sec_a_th1}. 
Theorem 1 proves that Eqs. \ref{eq5_} and \ref{eq6_} are equivalent and have the same representation ability. It is crucial to accelerate the model's speed with MHE since Eq. \ref{eq6_} can be employed to streamline the computational process.

Further, if the outputs $\tilde{\bm O}$ in Eq. \ref{eq5_} of the multi-head classifier with MHE are used to approximate the outputs $\bm O$ of the vanilla classifier with OHE in Eq. \ref{eq2}, we have the following corollary.
\begin{corollary}
  If $\bm{O} \approx \tilde{\bm O} = \bm{O}^1 \otimes \bm{O}^2 \otimes \cdots \otimes \bm{O}^H$, we have \label{co1}
  \begin{align}
    & \varLambda(\bm{O}) = \varLambda(\mathbb{I}_{\varLambda(\bm{O}^1)} \otimes \mathbb{I}_{\varLambda(\bm{O}^2)} \otimes \cdots \otimes  \mathbb{I}_{\varLambda(\bm{O}^H)}). \label{eq8_}
  \end{align}
\end{corollary}
The proof of Corollary \ref{co1} is given in Appendix \ref{sec_a_co1}. 
This corollary asserts that if the output of the vanilla classifier is decomposed into an approximation of the outputs of the multi-head classifier, then the predictions of the two classifiers are consistent.

\section{Implementations of MHE}
\label{sec_mhe_imp}

Now, we consider concretizing Eq. \ref{eq5_1} to render MHE applicable to various XLC tasks.
Specifically, MHP is applied to XSLC to o achieve multi-head parallel acceleration. MHC is used in XMLC to prevent confusion among multiple categories, and MHS is applied during model pre-training to efficiently extract features, as this task does not require a classifier.
% Given the differences in inference methods and application scenarios of XSLC, XMLC, and model pretraining, three MHE-based algorithms are designed: Multi-Head Product, Multi-Head Cascade, and Multi-Head Sampling.
Then, we provide a strategy to determine the number and length of the heads.

\subsection{Multi-Head Product (MHP)}
\label{sec4_1}

According to Corollary \ref{co1}, the output can be decomposed into the product of the heads, which paves the way for using MHP instead of the vanilla classifier to train the model.

As shown in Fig. \ref{fig3}a, during training, the global label $Y_i$ needs to be assigned to each head to calculate the loss locally. Thus, we first perform OHE on $Y_i$, then reshape it into an $H$-order tensor $\mathcal{Y}_i^{1,...,H}$ according to the length of the heads. Finally, $\mathcal{Y}_i^{1,...,H}$ is decomposed into local labels $\{Y_i^h\}_{h=1}^H$ on each head.
Since the decomposition of one-hot encoded $Y_i$ depends solely on the number and order of the heads, it can be recursively calculated as
\begin{equation}
  Y_i^h= \label{eq14_}
  \begin{cases}
    \lfloor \frac{Y_i}{\prod_{j=2}^H |\bm{O}^j|} \rfloor,  & h=1   \\ %
    \lfloor \frac{Y_i-\sum_{k=1}^{h-1} Y_i^k \prod_{j=k+1}^H |\bm{O}^j|}{\prod_{j=h+1}^H |\bm{O}^j|} \rfloor, & 2 \le h \le H-1 \\
    Y_i - \sum_{k=1}^{h-1} Y_i^k \prod_{j=k+1}^H |\bm{O}^j|,  & h=H
  \end{cases}
\end{equation}
where $j$ and $k$ are the indexes of the classification heads. 

During testing, the global prediction must be recovered from the local predictions. As shown in Fig. \ref{fig3}a, we first perform $\mathbb{I}_{\varLambda}$ on each head to obtain the locally predicted label. Then, the global prediction $\tilde{Y}_i$ is obtained by performing the product on each head and applying Argmax on the final outputs. To speed up this process, according to Theorem \ref{th1}, we calculate $\tilde{Y}_i$ from the local predictions and the length of the subsequent heads, as
\begin{equation}
  \tilde{Y}_i = \sum_{k=1}^{H-1} \varLambda(\bm{O}^k) \prod_{j=k+1}^H |\bm{O}^j| + \varLambda(\bm{O}^H).  \label{eq15_}
\end{equation}
The Algorithm for MHP is given in Appendix \ref{sec_af1}. It can be used in many XSLC tasks, such as image classification, face recognition, etc.

\subsection{Multi-Head Cascade (MHC)}
\label{sec4_2}

For XMLC, each sample $\bm{X}_i$ corresponds to multiple labels $\bar{\bm{Y}}_i \in \{0,1\}^{C}$, so the outputs of the classifier need to perform multi-hot encoding and Top-$K$ selection as $\tilde{Y}_i = \text{Top-}K(\bar{\bm{O}})$.
MHP cannot be adopted directly in XMLC. 
This arises from the fact that each head in MHP predicts only a single label. If utilized for multi-label predictions, it will result in a mismatch when computing the product of the local predictions.
To address this mismatch of MHP in multi-label scenarios, MHC is proposed, which cascades multiple heads for model training and testing. 

As shown in Fig. \ref{fig3}b, during training, the label decomposition process of MHC is the same as that of MHP. During testing, the top $K$ activations of the outputs are selected. Then, the local predictions of this head are obtained through a predefined candidate set $\mathbb{C}^1$, which is adopted to represent the label set of the subsequent heads to facilitate retrieval and reduce computation. 
The final outputs $\tilde{\bm {O}}^h$ of the $h$\text{-}th head are obtained by the product of embedded $\tilde{\bm{Y}}^{h-1}$ and current output $\bm{O}^h$. Then, $\tilde{\bm {Y}}^h$ is selected from $\mathbb{C}^h$ according to the top $K$ activations of $\tilde{\bm {O}}^h$. This process is repeated until the labels of $\tilde{\bm {Y}}^H$ are obtained as
\begin{equation}{\label{eq16_}}
  \begin{cases}
    \tilde{\bm{Y}}^1 = \mathbb{C}_{[1,...,i_{2}]}^{(i_1,|\bm{O}^{2}|)}[Top\text{-}K(\varphi(\bm{O}^1))] ,  & h=1   \\ %
    \tilde{\bm O}^{h} = \bm{O}^h {\mathbb{E}^h}^T(\tilde{\bm{Y}}^{h-1}),  & 2 \le h < H \\
    \tilde{\bm{Y}}^h = \mathbb{C}_{[1,...,i_{h+1}]}^{(i_h,|O^{h+1}|)} [\tilde{\bm{Y}}^{h-1}[Top\text{-}K(\varphi(\tilde{\bm{O}}^{h}))]], & 2 \le h < H  \\
    \tilde{\bm{Y}}^h = \tilde{\bm{Y}}^{h-1}[Top\text{-}K(\varphi(\tilde{\bm{O}}^{h}))] ,  & h=H
  \end{cases}
\end{equation}
where $\ i_h = \prod_{j=1}^h |\bm{O}^j|$, $\mathbb{E}^h$ is the embedding layer of the $h$\text{-}th head, and $\mathbb{C}_{[1,...,i_{h+1}]}^{(i_h,|\bm{O}^{h+1}|)}$ is the index matrix with elements from $1$ to $i_{h+1}$ and  shape $(i_h, |\bm{O}^{h+1}|)$. Eq. \ref{eq16_} shows that MHC is a coarse-to-fine hierarchical prediction method, which sequentially selects Top-$K$ candidate labels from the previous head. Note that MHC only depends on Eq. \ref{eq14_} for label decomposition and does not require preprocessing techniques like HLT or label clustering. The Algorithm for MHC is given in Appendix \ref{sec_af2}.

\subsection{Multi-Head Sampling (MHS)}
\label{sec4_3}

For the model pretraining tasks, the vanilla classifier is discarded when the training is completed, and only the features $\bm F$ extracted by the model are adopted for finetuning on downstream tasks. Therefore, all parameters of the weights in the classifier should be trained to extract more discriminative features, but training all parameters of the weights is computationally expensive. Therefore, MHS is proposed to update the model parameters by selecting the heads where ground truth labels are located.

As shown in Fig. \ref{fig3}c, MHS divides the vanilla classifier into $H$ groups equally, so that $\bm{O} = \sum_h^H |\bm{O}^h|$. During training, the head where label $Y_i$ is located is selected for model training, which is called the positive head. 
Certainly, we can also randomly select several negative heads to train the model together, thereby enriching the model with more negative sample information.
The forward process of MHS for ${\bm O}^{h}$ can be formulated as
\begin{subequations}{\label{eq15}}
  \begin{align}
    & \bm{O}^h =  \bm{O}^h \cup \{\bm{O}^j\} = \mathcal{W}^h\bm{F} \cup \{\mathcal{W}^j\}\bm{F},  \label{eq15a} \\
    & \bm{Y}^h = \bm{Y}^h \cup \{0\}  = \bm{Y}[|\bm{O}^{h-1}|:|\bm{O}^{h}|] \cup \{\bm{0}\},  \label{eq15b}
  \end{align}
\end{subequations}%\footnotetext[1]{}
where $\{\bm{O}^j\}$ and $\{\mathcal{W}^j\}$ indicate the outputs and weights set of the negative heads, respectively, and $\cup$ represents the concatenation operation. Eq. \ref{eq15b} indicates that $\bm{Y}^h$ is padded with $\bm 0$s to align the length of $\bm{O}^h$, where $|\bm{O}^h|=0$ when $h=0$.

The method in Eq. \ref{eq15} can be denoted as MHS-$S$, where $S$ is the number of selected heads. Our experiments show that MHS-$1$ (only the positive head) achieves quite good performance on model pretraining. For $S=2$, MHS approximates or outperforms the vanilla classifier. To speed up MHS, the heads containing the other sample labels in the same batch are selected as the negative heads. The Algorithm for MHS is given in Appendix \ref{sec_af3}.

\subsection{Label Decomposition Principle}
\label{sec4_4}

Thus far, we have introduced three MHE algorithms, the implementation of which is contingent upon both the number and the length of the heads.
Therefore, in this subsection, we introduce the concepts of error accumulation and confusion degree to measure the impact of the number and length of the heads on the performance of the MHE-based algorithms.

{\bf The number of heads}: 
The approximation process of MHE with H heads can be expressed as
\begin{align} \label{R2Q1E_1} % 头的个数越少，逼近误差越小；
  {\bm O} & \approx {\bm O}^1 \otimes \tilde{\bm O}^2 \approx {\bm O}^1 \otimes \underbrace{{\bm O}^2 \otimes \tilde{{\bm O}^3}}_{\approx \tilde{\bm O}^2} \notag \\
    & \approx {\bm O}^1 \otimes {\bm O}^2 \otimes \underbrace{\cdots \otimes {\bm O}^H}_{\approx \tilde{\bm O}^{H-1}}. 
\end{align}
As shown in Eq. \ref{R2Q1E_1}, adding a head is equivalent to accumulating one more time error.
Although increasing the number of heads will significantly reduce the parameters and calculations of the classifier, it will also cause greater cumulative errors. 
Thus, as long as the computing resources and running speed permit, the number of classification heads should be minimized. 

{\bf The length of heads}: 
The confusion degree is a measure of mismatches caused by shared components when MHE is adopted to approximate the original label space. It is proportional to the approximation error as
\begin{equation} \label{R2Q1E_CF1}
  D = \mathop{max}\limits_{\pi({\bm O}^1, \cdots, {\bm O}^H)} \left ( \prod_{h=2}^H \frac{\prod_{k=h}^H |{\bm O}^k|}{|{\bm O}^{k-1}|}\right ), \ \ (H \ge 2),
\end{equation}
where $\pi$ is an arrangement strategy of the heads.
The value of $D$ is expected to be as small as possible. 
Since $\pi$ relies on the specific decomposition process, we analyze the detailed confusion degrees for different algorithms of MHE. 

For MHP, the confusion degree is irrelevant to the arrangement of the heads because they are parallel and need to be combined. That is, $max$ in Eq. \ref{R2Q1E_CF1} can be removed when the length of the heads is in ascending order.
Therefore, we conclude that the length of each head in MHP should be as consistent as possible to minimize $D$, i.e., $|{\bm O}^h| \approx \sqrt[H]{C}$. 

For MHC, since heads are sequentially cascaded, we can choose a better strategy $\pi$ to minimize $D$. Obviously, when $\pi$ is in descending order ($|{\bm O}^1| \ge \cdots \ge {\bm O}^H$), $D$ is minimized. 

For MHS, those multiple heads are interrelated and need to be combined (irrelevant to the $max$ operation). That is, we can choose the same strategy as for MHC to minimize $D$.

\section{Representation Ability of MHE}
\label{sec_theorem_dis}

% \subsection{Representation Ability of MHE}
% \label{sec_rep_mhe}

As Corollary \ref{co1} proves, the essence of MHE is a low-rank approximation method that approximates high-order extreme labels through the product of multiple first-order heads.
Thus, one might inquire: \emph{Does MHE guarantee sufficiently robust performance in classification problems?}

\begin{figure}
  \centering
  \centerline{\includegraphics[width=0.97\columnwidth]{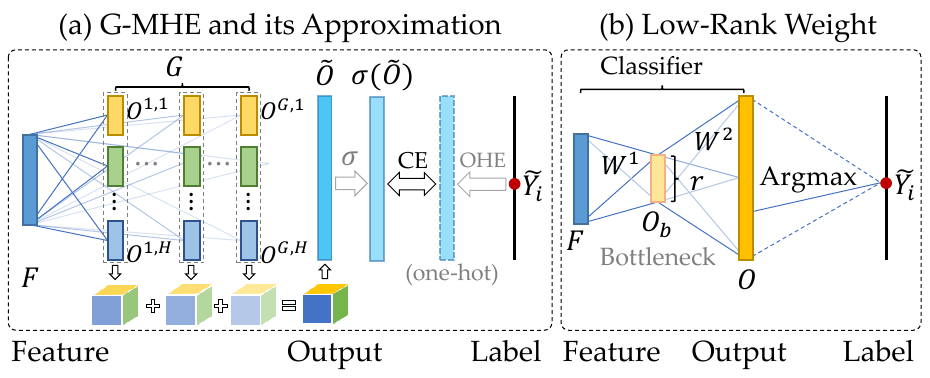}}
  \caption{The low-rank approximation ability of the classifier. (a) $G$ groups of multi-head classifier using MHE (G-MHE). (b) A bottleneck layer is added to the origin classifier to achieve the low-rank property of $\mathcal{W}$.}
  \label{fig1_5}
\end{figure}

\begin{figure*}
  \centering
  \centerline{\includegraphics[width=0.9\textwidth]{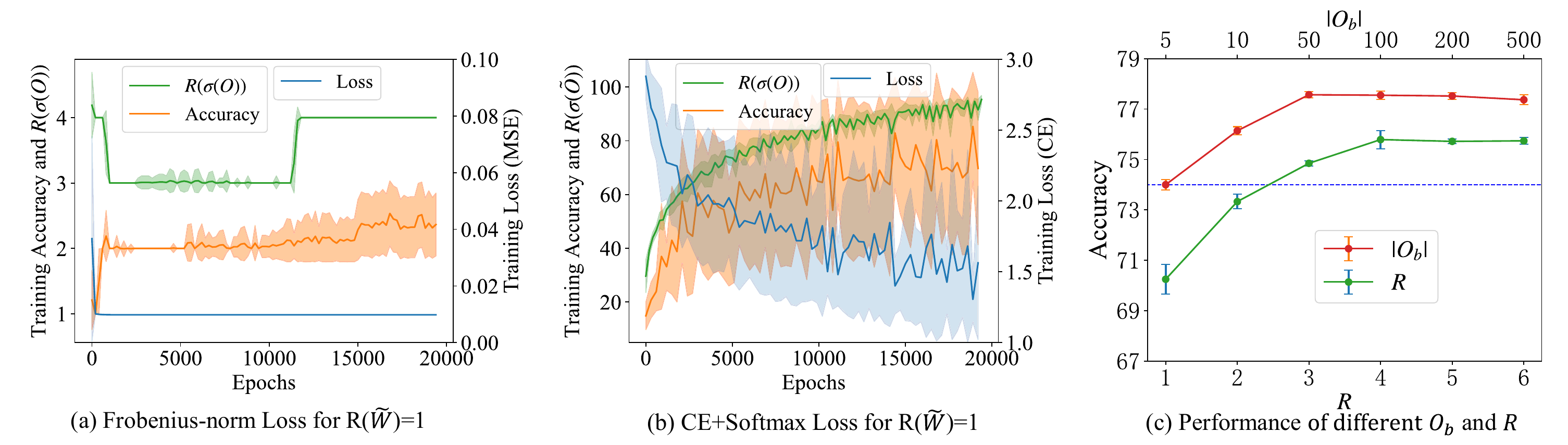}}
  \caption{Experiments with different loss functions and $R(\tilde{W})$. (a, b) The performance of two-layer linear networks on random samples generated from a Gaussian distribution. (c) The performance of ResNet-18 on the CIFAR-100 dataset.}
  \label{fig2}
 \end{figure*}

To answer the above question, we generalize MHE to a more general low-rank approximation problem, extending from the Frobenius-norm metric to Cross-Entropy.
As shown in Fig. \ref{fig1_5}a, if there are $G$ groups of multi-heads with MHE, each of which forms an $H$-order tensor. Then, all these tensors are added to obtain the final output
\begin{subequations}{\label{eq9_}}
  \begin{align}
    \mathbb{I}_{Y_i} & \approx \sigma(\tilde{\bm O})  = \sigma(\sum_g^G \bm{O}_g^1 \otimes \bm{O}_g^2 \otimes \cdots \otimes \bm{O}_g^H) \label{eq9a_} \\
    & = \sigma(\sum_g^G (\mathcal{W}_g^1F) \otimes (\mathcal{W}_g^2F) \otimes \cdots \otimes (\mathcal{W}_g^GF)),
    \label{eq9b_} 
  \end{align}
  \end{subequations}
where $g$ is the index of the groups. In fact, Eq. \ref{eq9_} is the CP decomposition of a tensor, with $G$ being its rank. This factorizes the tensor into a sum of component rank-one tensors. Theoretically, other tensor decomposition methods \cite{hitchcock1927expression, tucker1966some}
can also be used to approximate $\bm O$ when it is viewed as a vectorized high-order tensor. 

\subsection{Low-Rank Approximation with Frobenius-norm}

Equation \ref{eq9_} illustrates that the low-rank approximation of the output $\tilde{\bm O}$ is essentially to restrict the rank of its weights. Therefore, we study the impact of low-rank weights on the performance of the classifier. To constrain the rank of $\mathcal{W}$, a linear bottleneck layer $\bm{O}_b$ is added between the feature layer $\bm F$ and the output layer $\bm O$, as shown in Fig. \ref{fig1_5}b. Let the weight between $\bm F$ and $\bm {O}_b$ be denoted by $\mathcal{W}_1$, and the weight between $\bm{O}_b$ and $\bm O$ be denoted by $\mathcal{W}_2$. We have
\begin{align}
  & \bm {O} = \mathcal{W}_2\mathcal{W}_1 \bm{F} + \bm{B} =  \tilde{\mathcal{W}} \bm{F} + \bm{B}, \label{eq10_} \\
  & s.t. \ R(\tilde{\mathcal{W}}) = r \le min(|\bm{F}|,|\bm{O}_b|), \nonumber
\end{align}
where $\tilde{\mathcal{W}} = \mathcal{W}_2\mathcal{W}_1$ and $R(\cdot)$ is the rank of a matrix. If $\tilde{\mathcal{W}}$ is optimized by the Frobenius-norm loss, we have
\begin{equation}
  min \ L = \frac{1}{2} ||\mathbb{I}_{Y_i}-\bm{O}||_F^2 = ||\mathbb{I}_{Y_i}-\tilde{\mathcal{W}} \bm{F}||_F^2. \label{eq11_}
\end{equation}
Eq. \ref{eq11_} is a low-rank approximation problem \cite{baldi1989neural, zhu2018global} % zhu2020global, nouiehed2022learning
that makes all elements of $\mathbb{I}_{Y_i}$ and $\bm O$ as close as possible. It yields a low-rank approximation $\mathop{Argmin}\limits_{R(\tilde{\mathcal{W}}) \le r} ||\mathcal{W}-\tilde{\mathcal{W}}||_F^2$. Further, we have the following theorem.
\begin{theorem}
  Assume that $\mathcal F$ is of full row rank, using Frobenius-norm as the loss function to train the linear neural networks in Eq. \ref{eq10_} will result in no spurious local minima, and every degenerated saddle point $\mathcal{W}$ is either a global minimum or a second-order saddle.
\label{th2}
\end{theorem}

The proof of Theorem \ref{th2} is given in Appendix \ref{sec_a_th2}. 
This theorem states that any local optimal solution $\tilde{\mathcal{W}}^*$ in Eq. \ref{eq10_} is a global optimal solution, which means that the optimal approximation to $\mathbb{I}_{Y_i}$ can be obtained through any $\tilde{\mathcal{W}}^*$. 
It is worth noting that the full row rank condition specified in Theorem \ref{th2} can be readily satisfied in XLC tasks. 
This is because the length of the features is much smaller than the number of categories, i.e., $|\mathcal{F}\mathcal{F}^T| = |\mathcal{F}|, s.t. |F| \ll C$. 

\subsection{Low-Rank Approximation with CE}

Further, if the loss of the low-rank approximation in Eq. \ref{eq10_} is generalized from the Frobenius-norm to CE with softmax, we will get a better approximation of $\mathbb{I}_{Y_i}$. This is because the Frobenius-norm metric in Eq. \ref{eq10_} is too strict for the classification problem \cite{golik2013cross}, i.e., the Frobenius-norm loss tends to approximate all elements, while the CE loss tends to select the largest element. Therefore, the low-rank approximation in Eq. \ref{eq10_} needs to be generalized to the CE loss, which is commonly used but rarely studied in the classification problem.

Different from the Frobenius-norm used in Eq. \ref{eq10_}, the nonlinear operation on the outputs will affect their representation ability. 
This is because the softmax (training) and non-differentiable Argmax (testing) can be approximated by
\begin{subequations}{\label{eq12_}}
  \begin{align}
    & \varLambda(\bm{O}_i) = \mathop{lim}\limits_{\epsilon \rightarrow 0} \varLambda (\sigma_{\epsilon}(\bm{O}_i)) =  \mathop{lim}\limits_{\epsilon \rightarrow 0} \varLambda \left(\frac{e^{\frac{\bm{O}_i}{\epsilon}}}{\sum_j{e^{\frac{\bm{O}_i}{\epsilon}}}}\right),
  \end{align}
\end{subequations}
where $\epsilon$ is the temperature of the softmax. Equation \ref{eq12_} shows that the Argmax operation used in testing is actually consistent with the softmax and CE operations used in training. That is, Eq. \ref{eq12_} is equivalent to CE with softmax, where softmax makes the gap among elements larger and CE selects the largest element. Therefore, we generalize the low-rank approximation problem from the Frobenius-norm loss to the CE loss in the following theorem.
\begin{theorem}
  When $\mathcal F$ is separable, training the two-layer linear networks in Eq. \ref{eq10_} using CE with softmax as the loss function can recover the same accuracy as the vanilla classifier $\mathcal{O} = \mathcal{W} \mathcal{F}$, as long as $R([{\tilde{\mathcal{W}} \atop \bm{B}}]) > 1$ is satisfied.
  \label{th3}
\end{theorem}

The proof of Theorem \ref{th3} is given in Appendix \ref{sec_a_th3}. Theorem \ref{th3} shows that the minimum value of $R(\tilde{\mathcal{W}})$ can be equal to $1$ when the bias $\bm B$ exists, which means that the performance gap between OHE and MHE is considerably small.
Meanwhile, Theorem \ref{th3} also implies that when deep neural networks overfit the data, their generalization is irrelevant to the semantic information of the labels. This means that label preprocessing techniques, e.g., HLT and label clustering, are not necessary since the low-rank approximation remains independent of label positioning.

To validate this theorem, we generate $N\times N$ Gaussian random samples, where $N=100$ and $|\bm{O}_b|=1$. As shown in Fig. \ref{fig2}a, the training accuracy and $R(\sigma(\bm{O}))$ do not increase with the epochs. However, in Fig. \ref{fig2}b, $R(\sigma(\bm{O}))$ is positively correlated with the training accuracy, and it approaches $100\%$ as epochs increases. 
Then, to validate the selectivity of CE with softmax, we conducted experiments on CIFAR-100 \cite{Krizhevsky2009Learning} using ResNet-18 \cite{He2016ResNet} and setting $|\bm{O}_b|$ to be different lengths. The results are shown in Fig. \ref{fig2}c. It is found that when $|\bm{O}_b|$ is appropriately set, the test accuracy of the model can be well guaranteed. 
Experiments presented in Section \ref{ssec_lrd} provide further substantiation of this claim.

Furthermore, when CE with softmax is used to train the model in Eq. \ref{eq10_}, the approximation error of the low-rank matrix $\tilde{\mathcal{W}}$ can be analyzed by the following theorem.
\begin{theorem}
  Let $\mathcal{W}^*$ be a local minimum of the model in Eq. \ref{eq10_}, and $\Delta = \tilde{\mathcal{W}}-\mathcal{W}^*$, training the two-layer linear networks in Eq. \ref{eq10_} using CE with softmax as the loss function, we have
  \begin{equation}
    E \le  \sum_j^{C} |e^{\Delta_j}-1|, \label{eq13_}
  \end{equation}
  where $E$ is the approximation error of $\sigma(\mathcal{O})$ to $\sigma(\mathcal{O}^*)$.
\label{th4}
\end{theorem}
The proof of Theorem \ref{th4} is given in Appendix \ref{sec_a_th4}. Theorem \ref{th4} states that the approximation error $E$ is related to both the number of classes $C$ and the rank of $\tilde{\mathcal{W}}$. It illustrates an important conclusion: when $\Delta_j > 0$,  $E$ decreases exponentially, and when $\Delta_j \rightarrow 0$, $E$ decreases linearly. 
This is consistent with the deep learning method, where the loss drops sharply at the beginning of training.

\section{Experiments}
\label{sec_exp}

Intensive experiments are conducted on XSLC, XMLC, and model pretraining to fully validate the effectiveness of the three proposed MHE-based algorithms to cope with CCOP. 
%During the experiments, the number of heads and their lengths in the MHE-based methods can be found in Appendix \ref{sec_ae_hyper}.

\begin{figure}
  \centering
  \includegraphics[width=\columnwidth]{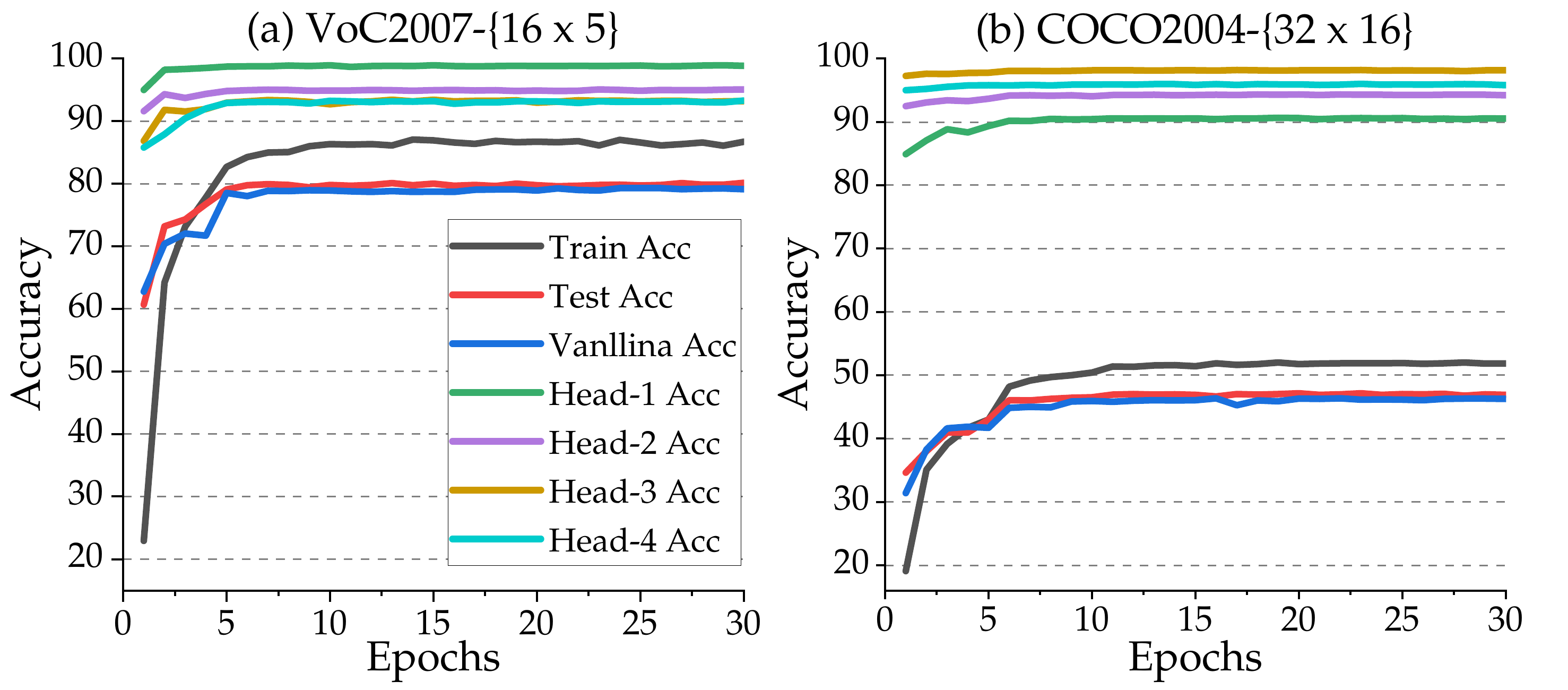}
  \caption{ \small 
  Testing and training accuracy of each classification head on the VOC2007 and COCO2014 datasets. Note that the label space of these datasets is transformed into the label powerset and is trained and tested on Resnet-101 using MHP. 
   }
   \label{fig_r2_11}
 \end{figure}

\begin{table}
  \centering
  \caption{\footnotesize Experiments on extreme version of VOC and COCO.}
  \label{tb1}
  \begin{threeparttable}
      \Large
      \resizebox{\columnwidth}{!}
      {
        \begin{tabular}{c|cc|cc}
          \toprule
          Dataset      & \multicolumn{2}{c|}{VOC}                                                & \multicolumn{2}{c}{COCO}            \\ \hline
          Vanilla       & Threshold-based                      & 79.82{\small $\pm 0.36$}        & Threshold-based                 & 46.62{\small $\pm 0.42$}  \\ \hline
          Labels        & \multicolumn{2}{c|}{Powerset=\{$2^{20}$\}}                             & \multicolumn{2}{c}{Powerset=\{$2^{80}$\}}                      \\ \hline
          AttentionXML  & HLT=\{$2^4;2^6;2^{10}$\}         & 77.74{\small $\pm 0.45$}            & HLT=\{$2^4;2^{16};2^{24};2^{32}$\}  & 46.06{\small $\pm 0.51$}  \\
          X-Transformer & LC=\{$2^{10}$;$2^{10}$\}         & 78.26{\small $\pm 0.45$}            &       --                            & --  \\
          LightXML      & LC=\{$2^{10}$;$2^{10}$\}         & 79.11{\small $\pm 0.37$}            &       --                            & --  \\
          MHP           & H =\{16$\times$5\}               & 80.13{\small $\pm 0.37$}            & H=\{32$\times$16\}                  & 47.16{\small $\pm 0.46$}  \\
          MHE           & H =\{16$\times$5\}               & 80.25{\small $\pm 0.35$}            & H=\{32$\times$16\}                  & 47.21{\small $\pm 0.44$}  \\
          MHS           & H =\{16$\times$5\}               & \bf{80.34}{\small $\pm 0.32$}       & H=\{32$\times$16\}                  & \bf{47.25}{\small $\pm 0.43$} \\ \bottomrule
        \end{tabular}
       }
  \scriptsize
  \begin{tablenotes}
      \item[*] The numbers in the set $\{\cdot \ ;\  \cdot \}$ indicate the setting of different XLC methods. \\`H' and `LC' represent the settings of heads and label clustering respectively.
  \end{tablenotes}
  \end{threeparttable}
\end{table}

\begin{table*}
  \Huge
  \centering
  \caption{Comparisons of MHC with other XMLC methods on the six public multi-label text datasets.}
  \label{tb3}
  \resizebox{0.97\textwidth}{!}
  {
    \begin{tabular}{cccccccccccccc}
      \toprule
      \multirow{2}{*}{Datasets}      & \multirow{2}{*}{P@K} & Annex & DiSM &Pfastre & Parabel & Bonsai & XT    & XML-  & XR-Li  & Attention & X-Trans & Light  & \multirow{2}{*}{MHC} \\
                                     &     & ML\cite{tagami2017annexml} & EC \cite{babbar2017dismec} & XML \cite{jain2016extreme} & \cite{prabhu2018parabel} & \cite{khandagale2020bonsai}  & \cite{wydmuch2018no}  & CNN \cite{liu2017deep} & near \cite{yu2022XRlinear} & XML \cite{you2019attentionxml}      & former \cite{chang2020taming}  & XML \cite{jiang2021lightxml}    &  \\ \midrule
      \multirow{3}{*}{Amazon-3M}     & P@1 & 49.30   & 47.34  & 43.83 & 42.20   & 48.45  & 42.20 & -     & 47.40 & 50.86        & 51.20         & -           & \bf{53.12}  \\
                                     & P@3 & 45.55   & 44.96 & 41.81 & 39.28   & 45.65  & 39.28 & -     & 44.15 & 48.00        & 47.81         & -           & \bf{49.54}  \\
                                     & P@5 & 43.11   & 42.80 & 40.09 & 37.24   & 43.49  & 37.24 & -     & 41.87 & 45.82        & 45.07         & -           & \bf{47.05}  \\ \midrule
      \multirow{3}{*}{Wiki-500K}     & P@1 & 64.22   & 70.21 & 56.25 & 68.70   & 69.26  & 65.17 & -     & 65.59 & 76.95        & 77.28         & 77.78       & \bf{79.74}    \\
                                     & P@3 & 43.15   & 50.57 & 37.32 & 49.57   & 49.80  & 46.32 & -     & 46.72 & 58.42        & 57.47         & 58.85       & \bf{60.90}    \\
                                     & P@5 & 32.79   & 39.68 & 28.16 & 38.64   & 38.83  & 36.15 & -     & 36.46 & 46.14        & 45.31         & 45.57       & \bf{47.21}    \\ \midrule   
      \multirow{3}{*}{Amazon-670K}   & P@1 & 42.09   & 44.78 & 36.84 & 44.91   & 45.58  & 42.54 & 33.41 & 43.38 & 47.58        & 48.07         & 49.10       & \bf{49.84}    \\
                                     & P@3 & 36.61   & 39.72 & 34.23 & 39.77   & 40.39  & 37.93 & 30.00 & 38.40 & 42.61        & 42.96         & 43.83       & \bf{44.42}    \\
                                     & P@5 & 32.75   & 36.17 & 32.09 & 35.98   & 36.60  & 34.63 & 27.42 & 34.77 & 38.92        & 39.12         & 39.85       & \bf{40.32}    \\ \midrule
      \multirow{3}{*}{Wiki10-31K}    & P@1 & 86.46   & 84.13 & 83.57 & 84.19   & 84.52  & 83.66 & 81.41 & 85.75 & 87.47        & 88.51         & 89.45       & \bf{89.51}    \\
                                     & P@3 & 74.28   & 74.72 & 68.61 & 72.46   & 73.76  & 73.28 & 66.23 & 75.79 & 78.48        & 78.71         & 78.96       & \bf{79.22}    \\
                                     & P@5 & 64.20   & 65.94 & 59.10 & 63.37   & 64.69  & 64.51 & 56.11 & 66.69 & 69.37        & 69.62         & 69.85       & \bf{70.35}    \\ \midrule
      \multirow{3}{*}{AmazonCat-13K} & P@1 & 93.54   & 93.81 & 91.75 & 93.02   & 92.98  & 92.50 & 93.26 & 94.64 & 95.92        & 96.70         & 96.77       & \bf{96.81}   \\
                                     & P@3 & 78.36   & 79.08 & 77.97 & 79.14   & 79.13  & 78.12 & 77.06 & 79.98 & 82.41        & 83.85         & 84.02       & \bf{84.07}  \\
                                     & P@5 & 63.30   & 64.06 & 63.68 & 64.51   & 64.46  & 63.51 & 61.40 & 64.79 & 67.31        & 68.58         & 68.70       & \bf{68.75}   \\ \midrule
      \multirow{3}{*}{Eurlex-4K}     & P@1 & 79.17   & 83.21 & 73.14 & 82.12   & 82.30  & 79.17 & 75.32 & 84.14 & 87.12        & 87.22         & 87.63       & \bf{87.94}    \\
                                     & P@3 & 66.80   & 70.39 & 60.16 & 68.91   & 69.55  & 66.80 & 60.14 & 72.05 & 73.99        & 75.12         & 75.89       & \bf{76.10}    \\
                                     & P@5 & 56.09   & 58.93 & 50.54 & 57.89   & 58.35  & 56.09 & 49.21 & 60.67 & 61.92        & 62.90         & 63.36       & \bf{63.78}    \\ \bottomrule
      \end{tabular}
  }
\end{table*}

\subsection{MHE-based Algorithms for XSLC}
\label{sec5_1}

To better illustrate the superiority of MHE on XLC tasks, we conduct experiments on the VOC2007 \cite{everingham2010pascal} and COCO2014 \cite{lin2014microsoft} datasets. The two datasets are well-known multi-label datasets. We transform their multi-labels into label powersets to build an XLC task.
The label space of the VOC2007 dataset is first transformed into the label powerset, whose size is $2^{20}$.
It should be noted that the vanilla classifier can no longer handle such a large label space with CCOP in it.
We compare the performance of the proposed MHE-based algorithms with three other SOTA XLC methods as shown in Table \ref{tb1}.
It can be found that MHE-based algorithms achieve advanced performance on the converted multi-label classification task.
It is worth noting that when the label space is too large (on the COCO dataset), X-Transformer \cite{chang2020taming} and LightXML \cite{jiang2021lightxml} fail to deal with CCOP because the label clustering process cannot be implemented.
Although AttentionXML \cite{you2019attentionxml} can handle XLC via HLT, its performance is seriously lower than that of the vanilla method since it involves complex optimizations at node splitting \cite{liu2021emerging}.

We also studied the prediction error for each head. 
As shown in Fig. \ref{fig_r2_11}a, the accuracy of each head exceeds 90\% on VOC2007, and one of them achieved 99.5\% (Head 1). The combined accuracy of the heads is about 80.13\%, which is competitive with the classical threshold-based method.
On dataset COCO2014, the size of the label powerset is $2^{80}$. 
% We adopt Resnet-101 with 16 heads to train and test this extreme label dataset. 
As shown in Fig. \ref{fig_r2_11}b, it shows 4 of the 16 heads due to the limited space. We find that the accuracy of each head exceeds 90\%, and the combined accuracy of the heads is lower. 
This is consistent with the characteristics of MHE, because the combined prediction result is the intersection of the prediction results of all heads.
However, as confirmed by the experimental results, the accumulation error of label decomposition is acceptable.
Especially for XLC tasks that traditional classifiers cannot handle, the advantages of the MHE-based algorithms are more obvious and prominent.

\begin{table*}
  \centering
  \caption{Single model comparisons of MHC with other Transformer-based XMLC methods.}
  \label{tb5}
  \begin{threeparttable}
    \begin{subtable}{0.97\textwidth}
      \resizebox{\textwidth}{!}
      {
        \begin{tabular}{c|ccc|ccc|cccc}
          \toprule
          Dataset & \multicolumn{3}{c|}{Eurlex-4K}            & \multicolumn{3}{c|}{AmazonCat-13K (MHC)}           & \multicolumn{4}{c}{Wiki10-31K} \\ \midrule
          \multirow{2}{*}{Method} & Attention & MHC         & MHC          & Attention & MHC       & MHC      & Attention & X-Trans & Light & MHC \\
                                  & XML       & \{86;46\}   & \{172;23\}  & XML  & \{155;86\} & \{310;43\}    & XML       & former  & XML     & \{499;62\} \\  \midrule
                          P@1     & 85.49     & 85.54       & \bf{86.00}  & 95.65     & 95.83     & \bf{95.94}  & 87.1      & 87.5    & 87.8      & \bf{89.40}           \\
                          P@3     & 73.08     & 72.97       & \bf{74.39}   & 81.93    & 82.10     & \bf{82.29}  & 77.8      & 77.2    & 77.3      & \bf{78.90}           \\
                          P@5     & 61.10     & 60.61       & \bf{62.05}   & \bf{66.90}     & 66.47     & 66.73  & 68.8      & 67.1    & 68.0      & \bf{70.25}    \\ \bottomrule
      \end{tabular}
      }
  \end{subtable}

    \begin{subtable}{0.97\textwidth}
      \vspace{1 ex}
      \resizebox{\textwidth}{!}
      {
        \begin{tabular}{c|cccc|cccc|c}
          \toprule
          Dataset & \multicolumn{4}{c|}{Wiki-500K}                                  & \multicolumn{4}{c|}{Amazon-670K}                 & Amazon-3M  \\ \midrule
          \multirow{2}{*}{Method}   & Attention & X-Trans & Light    & MHC          & Attention & X-Trans & Light   & MHC              & MHC               \\
                                    & XML       & former  & XML      & \{5630;89\}  & XML       & former  & XML     & \{8377;80\}      & \{8761;321\}       \\ \midrule
                            P@1      & 75.01     & 44.8    & 76.19   & \bf{78.38}   & 45.66     & 44.8    & 47.14   & \bf{47.82}       & \bf{50.26}            \\
                            P@3      & 56.49     & 40.1    & 57.22   & \bf{59.53}   & 40.67     & 40.1    & 42.02   & \bf{42.58}       & \bf{47.34}            \\
                            P@5      & 44.41     & 34.6    & 44.12   & \bf{46.04}   & 36.94     & 34.6    & 38.23   & \bf{38.57}       & \bf{44.58}            \\ \bottomrule
          \end{tabular}
        }
    \end{subtable}

    \footnotesize
    \begin{tablenotes}
      \item[*] Noting that the single model performance of other XMLC methods is not given on Amazon-3M dataset, and only MHC performance is given for future comparisons. The numbers in the set $\{\cdot \ ;\  \cdot \}$ indicate the length of each head.
    \end{tablenotes}
  \end{threeparttable}

\end{table*}

\subsection{MHC for XMLC}
\label{sec5_2}

For XMLC, MHC is validated on six different public benchmarking datasets, whose statistical information are shown in Table \ref{ap_tb2} of Appendix \ref{sec_ae2}. Precision@K (P@K) is utilized as the evaluation metric, which is widely adopted in XMLC. For all the experiments, AdamW \cite{kingma2014adam} with a learning rate of $1e\text{-}4$ is utilized as the optimizer for model training.
The model ensemble, dropout, \cite{srivastava2014dropout} and Stochastic Weight Averaging (SWA) \cite{izmailov2018averaging} techniques are adopted in many XMLC approaches recently. 
For example, in AttentionXML \cite{you2019attentionxml}, the authors propose to use three model ensembles and SWA to enhance performance. The following works \cite{chang2020taming, jiang2021lightxml} all adopt this setup, e.g., X-Transformer \cite{chang2020taming} adopts three SOTA pretrained Transformer-large-cased models to fine-tune, including Bert \cite{devlin2018bert}, Roberta \cite{liu2019roberta}, and XLNet \cite{yang2019xlnet}. LightXML \cite{jiang2021lightxml} also adopts these pretrained Transformer-based models for ensemble and uses SWA to alleviate overfitting. 
Thus, the proposed MHC adopts the same settings for fair comparisons. 
The other experimental setups can be found in Appendix \ref{sec_ae2}. 
% \footnote[1]{short for term frequency–inverse document frequency, is a numerical statistic that is intended to reflect how important a word is to a document in a collection or corpus.}

For comparison purposes, 11 SOTA methods are adopted as baselines. Comparisons are done on the six public multi-label text datasets.
The performances of MHC are shown in Tables \ref{tb3} and \ref{tb5}, which show that MHC achieves advanced performance in terms of different metrics on different datasets (as highlighted). The results based on model ensembles are shown in Table \ref{tb3}, showing that MHC outperforms those existing SOTA models on all datasets, which confirms that MHC is a simple and powerful method. 

Single model comparisons of MHC with recent Transformer-based XMLC methods are shown in Table \ref{tb5}. 
MHC outperforms many other methods, e.g., AttentionXML, X-Transformer and LightXML. It is worth noting that although offering such good performance, MHC makes no assumptions on the label space and does not adopt techniques such as HLT and label clustering for preprocessing. This means that additional and complex preprocessing techniques are not critical for XMLC tasks. 
Without HLT and label clustering techniques, MHC allows arbitrary partitioning of the label space, and assigns a longer main classification head (similar to a larger number of clusters), thus reducing the confusion of multiple labels and achieving improved performance. The ablation studies on the length of the classification heads also confirm the good performance achieved by MHC. 

In addition, the long-tailed label distribution is the most important problem in XMLC \cite{tail_label_8830456}. To verify the impact of this problem on the proposed method, we conduct experiments on XMLC datasets by trimming tail labels. To make the paper more readable, the related experimental results and discussions are put in Appendix \ref{sec_ae_longlbl}.

\begin{table*}
  \centering
  \caption{Comparisons of MHS with other model pretraining methods on face recognition tasks.}
  \label{tb6}
  \tiny
  \resizebox{0.95\textwidth}{!}
  {
    \begin{threeparttable}
    
    \begin{tabular}{cccccccccc}
      \midrule
      \multicolumn{2}{c}{Datasets}  & Method                                                        & IJB-B & IJB-C & LFW   & AgeDB & CALFW & CPLFW & CFP-FP  \\ \midrule
      \multirow{3}{*}{}             & \multicolumn{2}{c}{ArcFace(0.5)} \cite{deng2019arcface}       & 94.20 & 95.60 & 99.82 & -     & 90.57 & 84.00 & -      \\
                                    & \multicolumn{2}{c}{GroupFace} \cite{kim2020groupface}          & 94.90 & 96.30 & 99.85 & 98.28 & 96.20 & 93.17 & 98.63  \\
                                    & \multicolumn{2}{c}{CurricularFace} \cite{huang2020curricularface}      & 94.80 & 96.10 & 99.80 & 98.32 & 96.20 & 93.13 & 98.37  \\ \midrule
      \multirow{3}{*}{MS1MV2}       & \multicolumn{2}{c}{PartFC-0.1}  \cite{an2021partial}       & 94.40 & 95.80 & 99.82 & 98.13 & 96.15 & 92.95 & 98.48  \\
                                    & \multicolumn{2}{c}{ArcFace-Full} \cite{deng2019arcface, an2021partial}       & \bf{94.80} & \bf{96.20} & \bf{99.83} & 98.20 & 96.18 & 93.00 & 98.45  \\
                                    & \multicolumn{2}{c}{MHS-\{1994;43\}}       & 94.50    & 95.85 & \bf{99.83} & \bf{98.40} & \bf{96.20} & \bf{93.20} & \bf{98.64}  \\ \midrule
      \multirow{2}{*}{MS1MV3}       & \multicolumn{2}{c}{ArcFace-Full} \cite{deng2019arcface, an2021partial}       & -     & 95.02 & 99.85 & \bf{98.55} & -     & -     & 98.99  \\
                                    & \multicolumn{2}{c}{MHS-\{7187;13\}}       & 91.27 & \bf{95.06} & \bf{99.87} & \bf{98.55} & 96.17 & 93.45 & \bf{99.17}  \\ \midrule
      \multirow{3}{*}{CASIA}        & \multicolumn{2}{c}{PartFC-0.1} \cite{an2021partial} & 93.52 & 94.62 & 99.12 & \bf{93.23} & 92.57 & 85.72 & \bf{93.45}  \\
                                    & \multicolumn{2}{c}{ArcFace-Full} \cite{deng2019arcface, an2021partial}       & 93.14 & 94.31 & 99.13 & 92.98 & \bf{92.73} & 85.98 & 93.26  \\
                                    & \multicolumn{2}{c}{MHS-\{881;12\}}        & \bf{93.85} & \bf{95.06} & \bf{99.15} & 92.93 & 92.62 & \bf{86.28} & 93.26  \\ \midrule
    \end{tabular}
    \begin{tablenotes}
        \item[*] The 1:1 verification accuracies are given on the LFW, AGEDB-30, CALFW, CPLFW and CFP-FP datasets. TAR@FAR=1e-2 is reported on the IJB-B and IJB-C datasets when ResNet-18 is trained on CASIA dataset. TAR@FAR=1e-4 and 1e-5 are reported on the IJB-B and IJB-C datasets when ResNet-101 is trained on MS1MV2 and MS1MV3 datasets. The numbers in the set $\{\cdot \ ;\  \cdot \}$ indicate the length of each head.
    \end{tablenotes}
    \end{threeparttable}
  }
\end{table*}

\subsection{MHS for Model Pretraining}
\label{sec5_3}

For model preprocessing tasks that do not require the classification head for inference, we can use MHS to pretrain the model on a large dataset for better validation performance. The performance of model pretraining using MHS is tested on three different face recognition datasets, e.g., CASIA \cite{liu2015deep}, MS1MV2 \cite{guo2016ms, deng2019arcface}, and MS1MV3 \cite{guo2016ms, deng2020retinaface}. In the experiments, Arcface \cite{deng2019arcface} is adopted as the loss function, and the model is optimized using SGD with the Poly scheduler. All experiments are trained for $25$ epochs using ResNet-18 \cite{He2016ResNet} or ResNet-101 \cite{He2016ResNet} as the backbone network. The remaining experimental settings are included in Appendix \ref{sec_ae3}. The experimental results are shown in Table \ref{tb6}.

MHS outperforms many methods, e.g., ArcFace-Full \cite{deng2019arcface, an2021partial}, PartFC-0.1 \cite{an2021partial}, etc., and it achieves SOTA performance on multiple validation datasets, e.g., IJB-B \cite{whitelam2017iarpa}, IJB-C \cite{maze2018iarpa}, LFW \cite{huang2008labeled}, AGEDB-30 \cite{moschoglou2017agedb}, CALFW \cite{zheng2017cross}, CPLFW \cite{zheng2018cross}, and CFP-FP \cite{sengupta2016frontal}. After training on the dataset CASIA, MHS outperforms other methods in $4$ out of $7$ validation datasets. MHS also exceeds more than $4$ validation metrics on the MS1MV2 and MS1MV3 datasets compared to other methods, validating that MHS is an effective method to replace the vanilla classifier for model pretraining.

\subsection{Scalability of MHE}
\label{sec5_4}

%%% 这个地方得改。。。

To demonstrate that MHE can be easily extended to other XLC tasks, we verify the scalability of MHE on the Neural Machine Translation (NMT) tasks.
For NMT tasks, the classifier needs to represent and predict the probabilities of all tokens. In this subsection, we explore the possibility of using the MHC and MHS algorithms to replace the vanilla classifier. The OPUS-MT \cite{TiedemannThottingalEAMT2020} model is utilized to perform experiments on the ro-en and de-en datasets in WMT16 \cite{elliottEtAl2016VL16}, and the preprocessing part of OPUS-MT is removed to adapt to MHC. All experiments are fine-tuned for $3$ epochs using AdamW with a learning rate of $5e\text{-}5$.  The experimental results are included in Fig. \ref{fig4}c.
See Appendix \ref{sec_ae4} for the other experimental settings.

From Fig. \ref{fig4}c, it is seen that both MHC and MHS methods can not only achieve competitive performance but also provide the possibility to further expand the size of the vocabulary, thereby alleviating the out-of-vocabulary problem. The difference is that MHC can accelerate the language model in both training and inference phases, but the BLEU score is slightly lower than that of the original model. This requires additional alignment techniques to further improve the model's performance. We leave this for future work.

\begin{figure*}%[!ht]
  \centering
  \centerline{\includegraphics[width=0.9\textwidth]{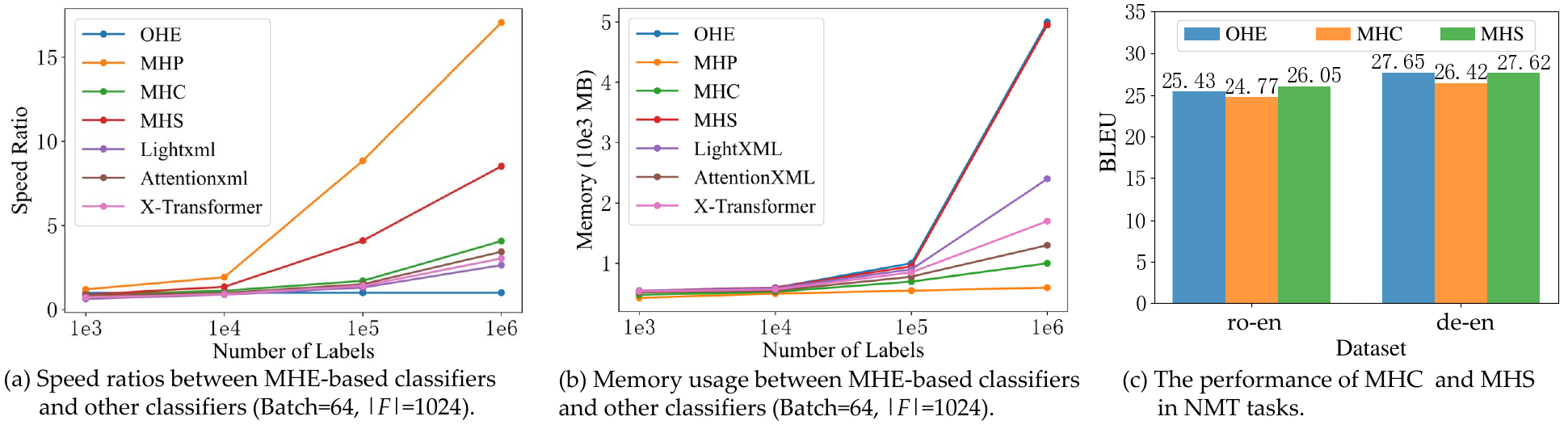}}
  \caption{The speed and performance comparisons of MHE-based algorithms.}
  \label{fig4}
\end{figure*}

\subsection{Time and Memory Consumptions}

We expanded the experimental comparisons to include validation of runtime and memory consumption, which are shown in Figs. \ref{fig4}a and \ref{fig4}b.
More information about model size can refer to the relevant backbone models \cite{He2016ResNet, devlin2018bert, liu2019roberta, yang2019xlnet, TiedemannThottingalEAMT2020}.
Since no preprocessing technology is used, MHE-based algorithms have advantages in speed and memory usage over other advanced XLC methods. 
Specifically, MHP has the fewest parameters and the fastest speed. As shown in Fig. \ref{fig4}a, when the number of labels reaches 1 million, MHP is 17$\times$ faster than OHE. Meanwhile, MHE and MHS are 4$\times$ and 8$\times$ faster than OHE, respectively. 
In terms of memory usage in Fig. \ref{fig4}b, MHP exhibits the lowest memory consumption, saving 4 GB of memory compared to that of the OHE method.
It is worth noting that MHS adopted for model pretraining has the same memory usage as OHE because it does not adopt parameter sharing. 
However, MHS is still 8$\times$ faster than OHE, and the performance of MHS is better than OHE, as shown in Table \ref{tb_times}.
These experimental results demonstrate that the MHE-based algorithms offer advantages in terms of both runtime and memory usage.

\begin{table}
  \centering
  \caption{Comparing runtime of different XLC methods.}
  \label{tb_times}
  \resizebox{0.95\columnwidth}{!}
  { \large
    \begin{threeparttable}
      \begin{tabular}{cccccccc}
        \toprule
        Dataset       & \begin{tabular}[c]{@{}l@{}}Attention\\ XML-3\end{tabular} & \begin{tabular}[c]{@{}l@{}}X-Trans\\ former-9\end{tabular} & \begin{tabular}[c]{@{}l@{}}Light\\ XML-3\end{tabular}   & MHC-3   & MHC-3 (parallel) \\
        \midrule
        Eurlex-4K     & 0.9            & 7.5             & 16.9        & \bf{0.9}             & \bf{0.6}     \\
        Wiki10-31K    & 1.5            & 14.1            & 26.9        & \bf{1.3}             & \bf{0.9}     \\
        AmazonCat-13K & 24.3           & 147.6           & 310.6       & \bf{21.4}           & \bf{9.7}     \\
        Wiki-500K     & 37.6           & 557.1           & 271.3       & \bf{31.5}           & \bf{31.3}     \\
        Amazon-670K   & 24.2           & 514.8           & 159.0       & \bf{20.7}           & \bf{8.5}   \\
        Amazon-3M     & 54.8           & 542.0           & -           & \bf{44.8}           & \bf{28.1}     \\
        \bottomrule
      \end{tabular}
      \small
      \begin{tablenotes}
        \item[*] The number following the model shows the number of ensembles adopted. The runtime is measured in hours on Tesla V100 GPU. Note that MHC (parallel) is trained on 8 Tesla V100s.
      \end{tablenotes}
    \end{threeparttable}
  }
\end{table}

\subsection{Ablation Studies of Label Decomposition Methods}
\label{ssec_lrd}

To further validate the conclusion implied in Theorem \ref{th3}, which states that the model generalization becomes irrelevant to the semantics of the labels when they overfit the data, we conduct ablation studies of label decomposition on model generalization.
It is known that the core of the preprocessing techniques is to perform semantic clustering on extreme labels and divide them into several tractable local labels.
Thus, we compare the performance of models utilizing label clustering (LC) with those employing label random rearrangement and arbitrary decomposition (LRD).

\begin{figure}%[!ht]
  \centering
  \begin{minipage}{\columnwidth}
    \includegraphics[width=\columnwidth]{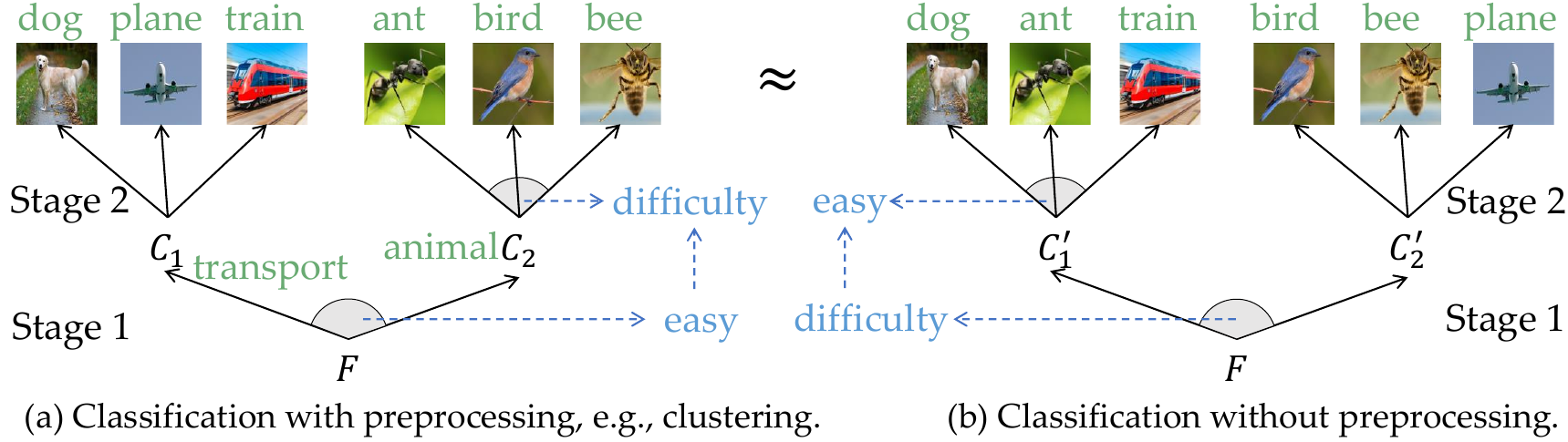}
    % \captionsetup{font={scriptsize},justification=justified}
    \caption{
      Comparisons between label decomposition with preprocessing (a) and without preprocessing (b) in XLC.  $F$ is the feature extracted from a model. $C_i$ is the $i$-th clusters, and $C'_i$ is the $i$-th random label sets. At each stage, there is a classifier that maps features to a certain set or category.
    }
  \label{fig_r1q1}
  \end{minipage}
\end{figure}

As shown in Fig. \ref{fig_r1q1}, label decomposition can be conceptualized as a multi-stage classification procedure, i.e., a given feature is initially assigned to a cluster, followed by the identification of a specific category within that cluster.
As shown in Fig. \ref{fig_r1q1}.a, preprocessing techniques can facilitate the initial stage of classification, e.g., distinguishing between the two categories of transports (dogs can be considered as a filled category)
and animals are proved to be easy due to their big differences.
However, in the second stage, classifying fine subcategories based on coarse features becomes difficult, e.g., it is more difficult to distinguish ants, birds, and bees from the animal cluster ($C_2$) than to distinguish birds, bees, and planes from the mixed cluster ($C'_2$), as shown in Fig. \ref{fig_r1q1}.b, in which no preprocessing technology is utilized. This suggests that without label preprocessing, the initial stage of LRD is relatively difficult, whereas the second stage is comparatively simpler than that of LC.
Please refer to Appendix \ref{subsec_a_lrd} for more experimental results on real datasets.

\subsection{Impact of Label Decomposition on Generalization}

\begin{figure}
  \centering
    \includegraphics[width=\columnwidth]{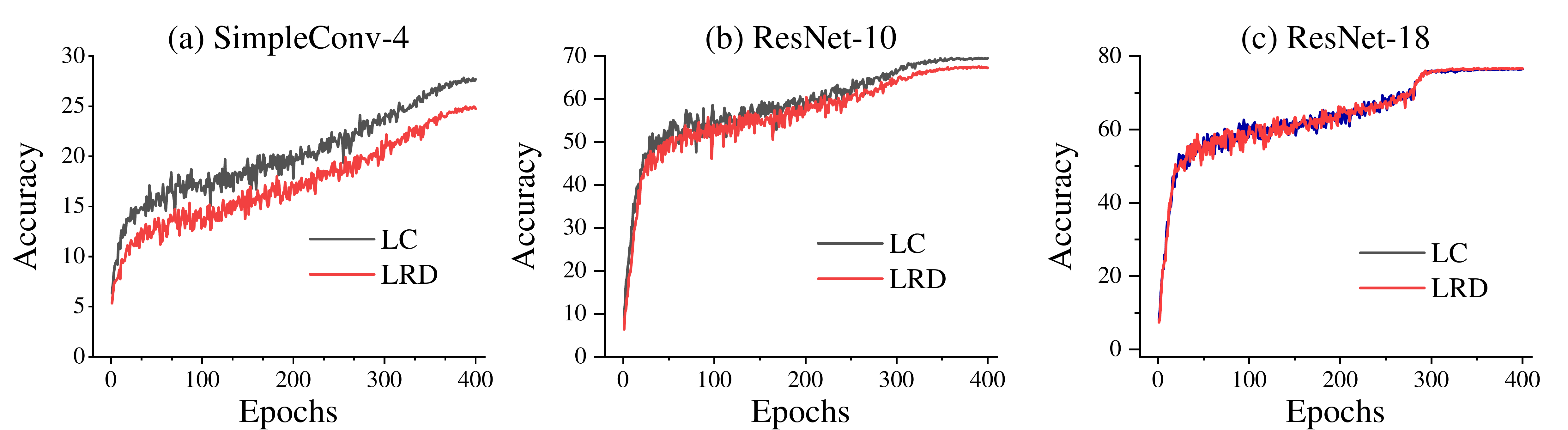}
    %\captionsetup{font={scriptsize},justification=justified}
    \caption{
      Ablation studies investigating the effects of label decomposition modes on model generalization. `LC' represents label clustering, and `LRD' represents label random rearrangement and then arbitrary decomposition. SimpleConv-4 (small) indicates underfitting, ResNet-10 (medium) indicates underfitting, and ResNet-18 (large) indicates overfitting.
    }
  \label{fig_r1q3}
\end{figure}

Furthermore, we compare three models of varying complexity to evaluate their generalization when configured using a carefully designed LC and a random LRD approach.
As shown in Fig. \ref{fig_r1q3}, 
when the low-complexity models underfit the data, there is a clear performance gap between LRD and LC: about 4\% in the small model (Fig. \ref{fig_r1q3}.a) and 2\% in the medium model (Fig. \ref{fig_r1q3}.b). 
This is due to the fact that a hierarchical classifier's performance in subsequent stages relies on the decision results made in earlier stages, especially when features extracted by low-complexity models exhibit reduced distinguishability.
This explains why the model with LRD falls behind the model with LC in situations involving low complexity.
However, this performance gap gradually diminishes as the model's complexity increases.
Eventually, when the model overfits the data, as shown in Fig. \ref{fig_r1q3}.c, the gap between LRD and LC vanishes.
It is noteworthy that the model's over-parameterization is readily achievable in practice, despite the high-complexity model (ResNet-18) employed here being small.
In addition, the experiments in this paper, including those in Appendix \ref{subsec_a_lrd}, support that the generalization performance of LC and LRD is consistent.

In summary, we find that LRD does not compromise generalization in overparameterized models when the number of clusters remains constant and the distribution of samples within them is approximately uniform. This strongly supports the claim implied in Theorem \ref{th3}.
% This claim is derived from the results of our LRD experimental validation conducted on numerous real-world datasets: The performance of the model with random LRD is essentially identical to that of the model with label processing.

\section{Discussion}
\label{sec_diss}

Here, we discuss the innovativeness of MHE by elucidating the distinctions between it and other methodologies that adopt multiple classifiers.

\begin{figure}
  \centering
  \centerline{\includegraphics[width=0.5\textwidth]{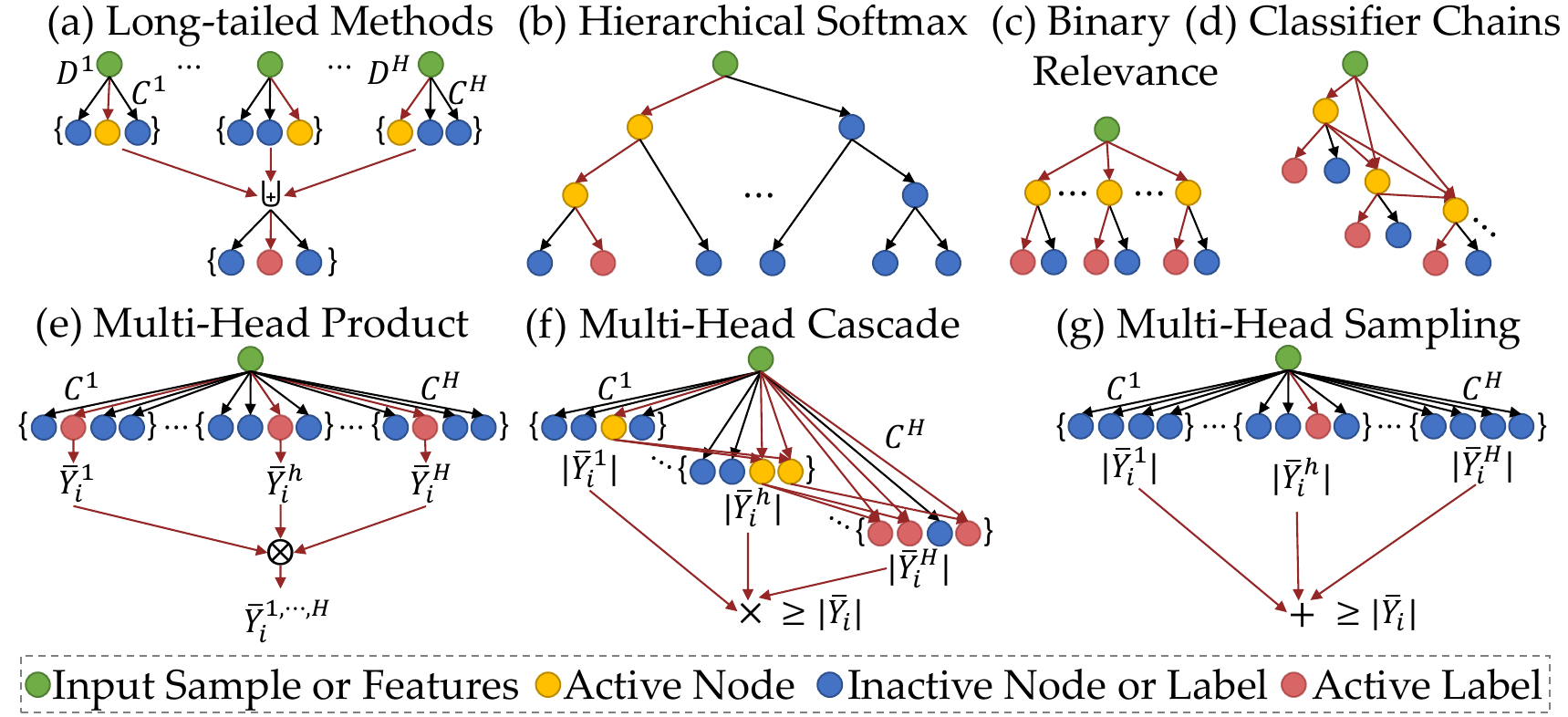}}
  \caption{Comparisons among algorithms using multiple classifiers. The symbol ``$\biguplus$'' denotes the aggregation operation and ``$\bigotimes$'' denotes the Kronecker product operation.}
  \label{fig_novelty}
\end{figure}

Several recent methods \cite{2020_ECCV_LFME, TPAMI_2022_ResLT} utilize multiple classifiers to address the long-tailed distribution problem. Specifically, the authors in \cite{2020_ECCV_LFME, TPAMI_2022_ResLT} split the dataset into balanced subsets and train an expert model on each subset. Then, multiple expert models (one model on one subset) are aggregated to obtain the final model, as shown in Fig. \ref{fig_novelty}a.
The long-tail methods are not applicable to solve CCOP, because the parameters of the classifier in the aggregated model have not been reduced.
Different form methods mentioned above, as shown in Fig. \ref{fig_novelty}(e-g), the proposed MHE-based algorithms can solve CCOP well by decomposing the hard-to-solve extreme labels into multiple easy-to-solve local labels, and combining local labels to obtain extreme labels through simple calculations.

There are many tree-based methods \cite{morin2005hierarchical, prabhu2014fastxml, prabhu2018parabel} using multiple classifiers for XLC tasks. These methods partition the label space through hierarchical branches.
For example, Hierarchical softmax \cite{morin2005hierarchical, mikolov2013distributed} adopts a Huffman tree to encode high-frequency words with short branches, as shown in Fig. \ref{fig_novelty}a.  
However, the huge label space greatly increases the depth and size of the tree, requiring traverse a deep path for low-frequency samples, making it unsuitable for XMLC tasks. 
Motivated by this, some HLT-based methods \cite{prabhu2014fastxml, prabhu2018parabel} have been proposed, but they involve complex optimizations at node splitting, making it difficult to obtain cheap and scalable tree structures \cite{liu2021emerging}. 
On the contrary, MHE-based algorithms have no preprocessing steps.
Therefore, the length of the classifier can be divided arbitrarily as long as the label space is fully mapped. 
% e.g., for MHP and MHC, $|\bar{Y}_i| \ge \prod_{h=1}^H{|\bar{Y}_i^h|}$ is satisfied, and for MHS, $|\bar{Y}_i| \ge \sum_{h=1}^H{|\bar{Y}_i^h|}$ is satisfied. 

%So they avoid the complex optimization problems that exist in HLT-based methods when doing nodes splitting \cite{liu2021emerging}. 

Several multi-label learning algorithms also adopt multiple classifiers to deal with the key challenge of the overwhelming size of the label powerset \cite{zhang2014review}. %e.g., if $|\bm{Y}|$=20, the number of possible output space would be $2^{20}$. 
Specifically, as shown in Fig. \ref{fig_novelty}c, the binary relevance algorithm \cite{boutell2004learning} decomposes the multi-label learning problem into $|\bm{Y}|$ independent binary classification problems. 
In Fig. \ref{fig_novelty}d, the classifier chains algorithm \cite{read2011classifier} transforms the multi-label learning problem into a chain of binary classification problems, where subsequent binary classifiers in the chain are built upon the predictions of preceding ones.
However, the number of classifiers in these algorithms is equal to the number of labels, which is not suitable for XLC tasks. 
Different from these algorithms, MHE-based algorithms are proposed to address CCOP by the combination of multi-head classifiers.
Therefore, the computational complexity of the MHES-based algorithm is significantly reduced, making it more flexible and better suited for XLC tasks.

\section{Conclusion}
\label{sec_con}

In this paper, we propose a Multi-Head Encoding (MHE) mechanism to cope with the challenge of CCOP that exists in XLC tasks.
MHE decomposes the extreme label into the product of multiple short local labels, and each head is trained and tested on these local labels, thus reducing the computational load geometrically.
For XLC tasks, e.g., XSLC, XMLC and model pre-training, three MHE-based algorithms are designed, including Multi-Head Product (MHP), Multi-Head Cascade (MHC), and Multi-Head Sampling (MHS). 
Experimental results show that the three MHE-based algorithms have achieved SOTA performance in the tasks to which they are applied and can greatly speed up model training and inference.
Furthermore, we conduct a theoretical analysis of the representation ability of MHE. It is proved that the performance gap between OHE and MHE is considerably small, and the label preprocessing techniques are unnecessary.

%extreme multi-label text classification and neural machine translation.

We believe that XLC is a natural extension of the traditional classification tasks, which allows us to deal with extreme labels, and is much more suitable for the real-world samples and practical applications. In turn, MHE-based algorithms designed for XLC can bring more novel solutions to many traditional tasks. For example, we can transform the regression task into an XLC task and use MHE-based algorithms to solve it. In reinforcement learning, MHE-based algorithms can provide accurate predictions for extreme state spaces when it is regarded as an XLC task.

\ifCLASSOPTIONcaptionsoff
  \newpage
\fi

\bibliographystyle{IEEEtran}
\bibliography{citations}

\clearpage
\setcounter{page}{1}

\appendices
\section{}

\subsection{Proof of Theorem \ref{th1}}
\label{sec_a_th1}

\begin{proof}
Substituting Eq. \ref{eq6_} into Eq. \ref{eq7_} and replacing $\varLambda$ with $\mathbb{I}_{\varLambda}$, we get
\begin{align}
   \mathbb{I}_{\varLambda(\tilde{\bm{O}})} =  \mathbb{I}_{\varLambda(\bm{O}^1)} \otimes 
  &\mathbb{I}_{\varLambda(\bm{O}^2)} \otimes \cdots \otimes  \mathbb{I}_{\varLambda(\bm{O}^H)}. \label{eq_ath1_1}
\end{align}
Here, we first consider the case of $H=2$ in Eq. \ref{eq_ath1_1}, and then generalize it to the case of $H>2$.

(a) $H=2$. Assume that the largest elements in $\tilde{\bm{O}}$, $\bm{O}^1$ and $\bm{O}^2$ are $o_i$, $o_k^1$ and $o_t^2$, respectively, there must have
\begin{align}{\label{eq_ath1_2}}
  & o_i = o_k^1o_t^2, \\
  s.t. \ \  &  o_i=max(\tilde{\bm{O}}), \nonumber \\
  & o_k^1= max(\bm{O}^1), \nonumber \\
  & o_t^2= max(\bm{O}^2). \nonumber
\end{align}
If Eq. \ref{eq_ath1_2} holds, it is implied that Eq. \ref{eq_ath1_1} also holds when $H=2$. Now we prove that Eq. \ref{eq_ath1_2} must hold.

1) If $o_k^1= max(\bm{O}^1)$ and $o_t^2= max(\bm{O}^2)$, there must be $o_i=max(\tilde{\bm{O}})$ such that Eq. \ref{eq_ath1_2} holds.

If there is another element $o_j<o_i$ in $\tilde{\bm{O}}$ that satisfies $o_j = o_k^1o_t^2$, then there exist two other elements $o_{k'}^1 \in \bm{O}^1, o_{t'}^2 \in \bm{O}^2$ that make $o_i = o_{k'}^1o_{t'}^2$ hold. Since $o_{k'}^1 \le o_{k}^1$ and $o_{t'}^2 \le o_t^2$, then $o_j \ge o_i$ must hold, which contradicts the assumption $o_j<o_i$.

2) If $o_i=max(\tilde{\bm{O}})$, there must be $o_k^1= max(\bm{O}^1)$ and $o_t^2= max(\bm{O}^2)$ such that Eq. \ref{eq_ath1_2} holds.

If there is another element $o_{k'}^1 \le o_{k}^1$ in $\bm{O}^1$ that satisfies $o_i = o_{k'}^1 o_t^2$, then there exists another element $o_j \in \tilde{\bm{O}}$ that make $o_j = o_k^1 o_t^2$ hold. Since $o_{k'}^1 \le o_{k}^1$, then  $o_j \ge o_i$ must hold, which contradicts the assumption $o_i=max(\tilde{\bm{O}})$. Similarly, this result also holds for $\bm{O}^2$.

(b) $H>2$. Using the associative property of the Kronecker product, Eq. \ref{eq_ath1_2} is equivalent to
\begin{align}{\label{eq_ath1_3}}
  \mathbb{I}_{\varLambda(\tilde{\bm{O}})} = \mathbb{I}_{\varLambda(\bm{O}^h)} \otimes \big( \mathbb{I}_{\varLambda(\bm{O}^{h-1})}  \otimes  
  \big( \cdots \otimes \left(\mathbb{I}_{\varLambda(\bm{O}^2)} \otimes \mathbb{I}_{\varLambda(\bm{O}^1)}\right)\big) \big).
\end{align}
Eq. \ref{eq_ath1_3} shows that $\bm{O}^1$ and $\bm{O}^2$ can be treated as a new vector for Kronecker product with $\bm{O}^3$, and the maximum value of the new vector can be guaranteed by Eq. \ref{eq_ath1_2}. This process continues until $\bm{O}^H$ is exploited, which means that Eq. \ref{eq7_} is hold.
\end{proof}

\subsection{Proof of Corollary \ref{co1}}
\label{sec_a_co1}

\begin{proof}
Replacing $\varLambda$ with $\mathbb{I}_{\varLambda}$ yields
\begin{align}
   \mathbb{I}_{\varLambda(\bm{O})} = \mathbb{I}_{\varLambda(\bm{O}^1)} \otimes
   \mathbb{I}_{\varLambda(\bm{O}^2)} \otimes \cdots \otimes  \mathbb{I}_{\varLambda(\bm{O}^H)}. \label{eq_d2}
\end{align}
Similar to Theorem \ref{th1}, we first consider the case of $H=2$ in Eq. \ref{eq_d2}, and then generalize it to the case of $H>2$.

(a) $H=2$. Assume that the largest element in $\bm{O}$ is $o_i$, there is an element $o_j \in \bm{O}, j\neq i$, satisfing that $o_i>o_j$. Let $o_k^1, o_t^2$ be any two elements of $\bm{O}^1$ and $\bm{O}^2$, respectively, when $H=2$, according to condition $ \bm{O} \approx \bm{O}^1 \otimes \bm{O}^2 $, we have
\begin{align}{\label{eq_d3}}
  & o_i = o_k^1o_t^2 + \epsilon_i, \\
   s.t. \ \ & o_k^1 = max(\bm{O}^1), \nonumber \\
  & o_t^2 = max(\bm{O}^2), \nonumber
\end{align}
where $\epsilon_i$ is the approximation error of $o_i$. Eq. \ref{eq_d3} shows that when $o_i$ is the maximum in $\bm{O}$, $o_k^1$ and $o_t^2$ are the maximum in $\bm{O}^1$ and $\bm{O}^2$, respectively. If there is an element $o_m^1$ in $\bm{O}^1$, satisfying that
\begin{align}{\label{eq_d4}}
  & o_j = o_m^1 o_t^2 + \epsilon_j, \\
  & s.t. \ \  o_m^1 > o_k^1, \nonumber
\end{align}
the conclusion in Eq. \ref{eq_d4} does not hold. Next, we will show that Eq. \ref{eq_d4} is impossible when $\bm{O}^1$ and $\bm{O}^2$ are the optimal approximation of $\bm{O}$. Note that the optimal approximation means that the approximation error $E = \sum_{k} |\epsilon_k|$ is minimal.

\begin{enumerate}[leftmargin=*]
\item  When $\epsilon_i < 0$, there is $\epsilon_j < \epsilon_i < 0$ due to $o_m^1 o_t^2 > o_k^1 o_t^2$ and $o_i > o_j$.  If $o_m^1$ in Eq. \ref{eq_d4} is replaced with $o_k^1$ in Eq. \ref{eq_d3}, $\epsilon_j$ can be made smaller, which contradicts the minimality of $E$.
\item When $\epsilon_i > 0$ and $\epsilon_j > 0$, there is $o_i - o_m^1 o_t^2< \epsilon_i $, which contradicts the minimality of $E$.
\item When $\epsilon_i > 0$ and $0 \ge \epsilon_j > -\epsilon_i$, there is $ -\epsilon_i < o_j - o_m^1 o_t^2 < 0$. Replacing $o_j$ with $o_i$ gives $o_i - o_m^1 o_t^2 < \epsilon_i$, which contradicts the minimality of $E$.
\item When $\epsilon_i > 0$ and $\epsilon_j \le -\epsilon_i$, it means that $o_j - o_k^1 o_t^2 < |\epsilon_j|$, which contradicts the minimality of $E$.
\end{enumerate}
Based on the above analysis, it must have $o_k^1 = max(\bm{O}^1)$. Similarly, it must have $o_t^2 = max(\bm{O}^2)$. Therefore, the case of $H=2$ in Eq. \ref{eq_d2} is proved.

(b) $H>2$. Using the associative property of the Kronecker product, Eq. \ref{eq_d2} is equivalent to
\begin{align}{\label{eq_d5}}
  \mathbb{I}_{\varLambda(\bm{O})} = \mathbb{I}_{\varLambda(\bm{O}^h)} \otimes \big( \mathbb{I}_{\varLambda(\bm{O}^{h-1})}  \otimes 
  \big( \cdots \otimes \left(\mathbb{I}_{\varLambda(\bm{O}^2)} \otimes \mathbb{I}_{\varLambda(\bm{O}^1)}\right)\big) \big).
\end{align}
Eq. \ref{eq_d5} shows that $\bm{O}^1$ and $\bm{O}^2$ can be treated as a new vector for Kronecker product with $\bm{O}^3$, and the maximum value of the new vector can be guaranteed by Eq. \ref{eq_d3}. This process continues until $\bm{O}^H$ is exploited, which means that Eq. \ref{eq8_} is hold.
\end{proof}

\section{Proof of Theorem \ref{th2}}
\label{sec_a_th2}

\begin{proof}

(a) Before proving Theorem \ref{th2}, we first prove a simplified form of it: the one-layer neural networks trained by the Frobenius-norm loss. As represented by Eq. \ref{eq10_} or Eq. \ref{eq11_}, it can be formalized as
\begin{align}{\label{eq_a1}}
  & \mathop{min}\limits_{\tilde{\mathcal{W}}} \  L(\tilde{\mathcal{W}}) = \frac{1}{2} || \tilde{\mathcal{W}}\mathcal{F} - \mathcal{Y} ||_F^2, \\
  & s.t. \ \ R(\tilde{\mathcal{W}})<r.  \nonumber
\end{align}
%Although it is studied in [A,B,C]. Here, we just give a brief proof of it
Next, we prove that every local minimum of Eq. \ref{eq_a1} is the global minimum.

Let $U_f\Sigma_fV_f^T$ be a singular value decomposition of $\mathcal{F}$. Let $\hat{U}_f \hat{\Sigma}_f \hat{V}_f^T$ be its reduced SVD, which is truncated according to the $R(\mathcal{F})$, we have
\begin{align}\label{eq_a2}
  \frac{1}{2}||\tilde{\mathcal{W}}\mathcal{F}-\mathcal{Y}||_F^2 &= \frac{1}{2}||\tilde{\mathcal{W}}U_f\Sigma_f - \mathcal{Y}V_f^T||_F^2 \nonumber \\
  & = \frac{1}{2} ||\tilde{\mathcal{W}}\hat{U}_f\hat{\Sigma}_f - \mathcal{Y}\hat{V}_{f}||_F^2  \nonumber \\
  & +  \frac{1}{2}|| \mathcal{Y}V_f ||_F^2 - \frac{1}{2}|| \mathcal{Y}\hat{V}_f ||_F^2.
\end{align}
The item $\frac{1}{2}|| \mathcal{Y}V_f^T ||_F^2 - \frac{1}{2}|| \mathcal{Y}\hat{V}_f^T ||_F^2$ in Eq. \ref{eq_a2} is constant, so it can be dropped and does not affect the minimization problem. Therefore, Eq. \ref{eq_a1} is equivalent to
\begin{align}{\label{eq_a3}}
  & \mathop{min}\limits_{\tilde{\mathcal{W}}} \ L(\tilde{\mathcal{W}}) = \frac{1}{2} ||\tilde{\mathcal{W}}\hat{U}_f\hat{\Sigma}_f - \mathcal{Y}\hat{V}_{f}||_F^2,  \\
  & s.t. \ \ R(\tilde{\mathcal{W}})<r. \nonumber
\end{align}
Let $\mathcal{Y}\hat{V}_{f} = U_Y\Sigma_YV_Y^T$ be the SVD of $\mathcal{Y}\hat{V}_{f}$, and $\hat{U}_Y\hat{\Sigma}_Y\hat{V}_Y^T$ be its reduced SVD, we have
\begin{align}{\label{eq_a4}}
  \mathop{min}\limits_{\tilde{\mathcal{W}}} \ L(\tilde{\mathcal{W}}) & = \frac{1}{2} ||\tilde{\mathcal{W}}\hat{U}_f\hat{\Sigma}_f - U_Y\Sigma_YV_Y^T||_F^2 \nonumber \\
  & = \frac{1}{2} ||\bar{U}_Y^T\tilde{\mathcal{W}}\hat{U}_f\hat{\Sigma}_f\bar{V}_Y - \Sigma_Y||_F^2,
  %= \frac{1}{2} || A - \Sigma_Y||_F^2
\end{align}
where $\bar{U}_Y$ and $\bar{V}_Y$ are obtained by padding zeros on the columns or rows of $U_Y$ and $V_Y$ when $R(\mathcal{Y}\hat{V}_f)< R(\tilde{\mathcal{W}}\hat{U}_f\hat{\Sigma}_f)$. Further, let $\mathcal{A} = \bar{U}_Y^T\tilde{\mathcal{W}}\hat{U}_f\hat{\Sigma}_f\bar{V}_Y$, we have
\begin{align}{\label{eq_a5}}
  & \mathop{min}\limits_{\tilde{\mathcal{W}}} \ L(\tilde{\mathcal{W}}) = \frac{1}{2} || \mathcal{A} - \Sigma_Y||_F^2, \\
  & s.t. \ \ R(\tilde{\mathcal{W}})<r. \nonumber
\end{align}
Let $\mathcal{A}^*$ denote the local minimum of Eq. \ref{eq_a5}, $\hat{U}_A\hat{\Sigma}_A\hat{V}_A^T$ be the reduced SVD of $\mathcal{A}^*$, and $\mathcal{P}_L = \hat{U}_A\hat{U}_A^T$ be the projection matrix of $\hat{U}_A$, $\mathcal{A}^*$ must satisfy
\begin{align}{\label{eq_a6}}
  & \mathop{min}\limits_{\tilde{\mathcal{W}}} \ L(\tilde{\mathcal{W}}) = \frac{1}{2} || \mathcal{A} - \Sigma_Y||_F^2, \\
  & s.t. \ \ \mathcal{P}_L\mathcal{T}=\mathcal{T}. \nonumber
\end{align}
Eq. \ref{eq_a6} is a convex problem, and its solution is the subset of Eq. \ref{eq_a5}. The minimum of Eq. \ref{eq_a6} is $\mathcal{A}^*=\mathcal{P}_L\Sigma_Y=\hat{U}_A\hat{U}_A^T\Sigma_Y$, which depends on $\mathcal{F}$, $Y$ and $R(\tilde{\mathcal{W}})$, regardless of the values of $\tilde{\mathcal{W}}$. That is, given $\mathcal{F}$, $\bm Y$ and $R(\tilde{\mathcal{W}})$, the minimum of the geometric landscape in Eq. \ref{eq_a1} is fixed. Therefore, every local minimum in Eq. \ref{eq_a1} is the global minimum.

Note that when $\mathcal{F}$ is full rank and $R(\tilde{\mathcal{W}})=min\{R(\mathcal{F}),R(\mathcal{Y})\}$, $\mathcal{P}_L$ is an identity matrix $\mathcal{I}$ ($\hat{U}_A=U_A$), and the loss function $L$ in Eq. \ref{eq_a1} is $0$. When $R(\tilde{\mathcal{W}})<min\{R(\mathcal{F}),R(\mathcal{Y})\}$, $\mathcal{P}_L$ is a $R(\tilde{\mathcal{W}}) \text{-}$block diagonal matrix, and the loss function $L(\tilde{\mathcal{W}})$ increases as $R(\tilde{\mathcal{W}})$ decreases. Further, when $R(\tilde{\mathcal{W}})<min\{R(\mathcal{F}),R(\mathcal{Y})\}$, the model in Eq. \ref{eq_a1} has many strict saddle points (the Hessian at any saddle point has a negative eigenvalue), which can be escaped by increasing the perturbation (increasing $R(\tilde{\mathcal{W}})$).

(b) Now, we prove that the model in Eq. \ref{eq10_} has no spurious local minimum, and every degenerated saddle point $\tilde{\mathcal{W}}^*=(\mathcal{W}_2^*,\mathcal{W}_1^*)$ is either a global minimum or a second-order saddle point. We repeat the objective function that
\begin{align}
  & \mathop{min}\limits_{\mathcal{W}_1, \mathcal{W}_2} \ L(\mathcal{W}_2,\mathcal{W}_1)
  = \frac{1}{2} || \mathcal{W}_2\mathcal{W}_1 \mathcal{F} - \mathcal{Y}||_F^2, \label{eq_a7} \\
  & s.t. \ \  R(\mathcal{W}_2\mathcal{W}_1) \le min\{|\mathcal{F}|,|\mathcal{O}_b|,|\mathcal{Y}|\}. \nonumber
\end{align}

When $R(\mathcal{W}_2\mathcal{W}_1) = min\{|\mathcal{F}|,|\mathcal{O}_b|,|\mathcal{Y}|\}$, $\mathcal{W}_2\mathcal{W}_1$ can be regraded as $\tilde{\mathcal{W}}$ in Eq. \ref{eq_a1}, and the proof is given in part (a). When $R(\mathcal{W}_2\mathcal{W}_1) < min\{|\mathcal{F}|,|\mathcal{O}_b|,|\mathcal{Y}|\}$, $\tilde{\mathcal{W}}^*$ is a degenerated critical point. According to the first-order prerequisite of Eq. \ref{eq_a7}, we have
\begin{align}
  & \nabla_{\mathcal{W}_1} L =   \mathcal{W}_2^T(\mathcal{W}_2\mathcal{W}_1\mathcal{F} - \mathcal{Y})\mathcal{F}^T, \label{eq_a8} \\
  & \nabla_{\mathcal{W}_2} L =  (\mathcal{W}_2\mathcal{W}_1\mathcal{F} - \mathcal{Y})\mathcal{F}^T\mathcal{W}_1^T. \nonumber
\end{align}
Let $\Delta^* = \mathcal{W}_2^*\mathcal{W}_1^*\mathcal{F} - \mathcal{Y}$, the degenerated critical point $\tilde{\mathcal{W}}^*$ must satisfy $\langle \Delta^*, \mathcal{F} \rangle \neq 0$ according to Eq. \ref{eq_a8}. Otherwise, $\tilde{\mathcal{W}}^*$ is the global minimum point  due to the convexity of $L$.
Suppose that there is a perturbation $\Delta{\tilde{\mathcal{W}}} = (\Delta{\mathcal{W}_2},\Delta{\mathcal{W}_1})$ near $\tilde{\mathcal{W}}^*$, the first-order optimality condition of Eq. \ref{eq_a7} can be obtained by
\begin{align}
  & \nabla L(\tilde{\mathcal{W}}^*) = \ \langle \Delta{\mathcal{W}_2}\mathcal{W}_1^*\mathcal{F} + \mathcal{W}_2^*\Delta{\mathcal{W}_1}\mathcal{F}, \Delta^* \rangle , % \ge 0
   \label{eq_a9}
\end{align}
and the second-order optimality condition of Eq. \ref{eq_a7} can be obtained by
\begin{align}
  & \nabla^2 L(\tilde{\mathcal{W}}^*) = ||\Delta{\mathcal{W}_2}\mathcal{W}_1^*\mathcal{F} + \mathcal{W}_2^*\Delta{\mathcal{W}_1}\mathcal{F}||_F^2 \nonumber \\
  & + 2 \langle\Delta{\mathcal{W}_2}\Delta{\mathcal{W}_1}F, \Delta^* \rangle . % \ge 0
   \label{eq_a10}
\end{align}
If $\tilde{\mathcal{W}}^*$ is a non-degenerate critical point, then $\nabla L(\tilde{\mathcal{W}}^*)=0$ in Eq. \ref{eq_a9} and $\nabla^2 L(\tilde{\mathcal{W}}^*)\ge 0$ in Eq. \ref{eq_a10} must be hold. Thus, we have
\begin{align}
   \nabla^2 L(\tilde{\mathcal{W}}^*)  = & ||\Delta{\mathcal{W}_2}\mathcal{W}_1^*\mathcal{F} + \mathcal{W}_2^*\Delta{\mathcal{W}_1}\mathcal{F}||_F^2||{\Delta^*}^T||_F^2 \nonumber \\
  & + 2 \langle\Delta{\mathcal{W}_2}\Delta{\mathcal{W}_1}\mathcal{F}, \Delta^* \rangle||{\Delta^*}^T||_F^2  \ge 0 \nonumber \\
  \Rightarrow & 2 \langle\Delta{\mathcal{W}_2}\Delta{\mathcal{W}_1}\mathcal{F}, \Delta^* \rangle||{\Delta^*}^T||_F^2  \ge 0 \nonumber \\
  \Rightarrow & \langle\Delta{\mathcal{W}_2}\Delta{\mathcal{W}_1}\mathcal{F}, \Delta^* \rangle \ge 0.
   \label{eq_a11}
\end{align}
Eq. \ref{eq_a11} shows that $\nabla^2 L(\tilde{\mathcal{W}}^*)\ge 0$ is not guaranteed because we can adjust the perturbation $\Delta{\mathcal{W}_2}$ and $\Delta{\mathcal{W}_1}$ to make $\langle\Delta{\mathcal{W}_2}\Delta{\mathcal{W}_1}F, \Delta^* \rangle < 0$, which contradicts the assumption that $\tilde{\mathcal{W}}^*$ is a non-degenerate critical point. When $\langle\Delta{\mathcal{W}_2}\Delta{\mathcal{W}_1}\mathcal{F}, \Delta^* \rangle = 0$, it means that $\langle \Delta^*, \mathcal{F} \rangle = 0$, which consistent with Eq. \ref{eq_a8}, indicating that $\tilde{\mathcal{W}}^*$ is the global minimum.
Therefore, every critical point $\tilde{\mathcal{W}}^*$ of the two-layer linear neural networks in Eq. \ref{eq_a7} is either a global minimum or a second-order saddle point.

\end{proof}

\section{Proof of Theorem \ref{th3}}
\label{sec_a_th3}
\begin{proof}
The objective function in Theorem \ref{th3} can be formulated as
\begin{align}{\label{eq_b1}}
& L = \sum_i^N \bm{Y}_i^T log(\sigma(\mathcal{W}_2\mathcal{W}_1\bm{F}_i+\bm{B})), \\
& s.t. \ \ 1< R([{\mathcal{W}_2\mathcal{W}_1 \atop \bm{B}}]) \le min\{|\mathcal{F}|,|\mathcal{O}_b|,|\mathcal{O}|\}, \nonumber
\end{align}
where $\mathcal{F} \in R^{|\mathcal{F}|\times N}, \mathcal{W}_1 \in R^{|\mathcal{O}_b| \times |\mathcal{F}|}, \mathcal{W}_2 \in R^{|\mathcal{O}|\times |\mathcal{O}_b|}, \bm{B} \in R^{|\mathcal{O}|}$. We now only prove the case of $R([{\mathcal{W}_2\mathcal{W}_1 \atop \bm{B}}])=2$, which can be naturally generalized to the case of $R([{\mathcal{W}_2\mathcal{W}_1 \atop \bm{B}}])>2$.

On the one hand, the single-layer linear network $\mathcal{O}=\mathcal{W}\mathcal{F}+\bm{B}$ in Eq. \ref{eq2} can be formulated as
\begin{align}{\label{eq_b1_1}}
  & \mathcal{O}^{[j]} = \mathcal{W}^j\mathcal{F}^{[j]} + \bm{B}^j,  \\
  & s.t. \ \ \mathcal{O}^{[j]} > max(\{ \mathcal{O}^{[k \neq j]} \}_{k=1}^C), \nonumber
\end{align}
where $j \in \{1,\cdots,C\}$, $\mathcal{W}^j \in R^{|\mathcal{F}| \times 1}$, the superscript $[j]$ represents the sample set whose label is $j$. Eq. \ref{eq_b1_1} shows that the probability of the $j$-th class can be maximized by adjusting only $\mathcal{W}_j$.

On the other hand, considering that $\mathcal{F}$ is separable, there are no two completely different samples corresponding to the same label. When $R([{\mathcal{W}_2\mathcal{W}_1 \atop \bm{B}}])=2$, it means that $|\mathcal{O}_b|=1$ and $R(\mathcal{W}_2\mathcal{W}_1)=1$ in Eq. \ref{eq10_}. Similar to Eq. \ref{eq_b1_1}, the projection results $O_b=W_1F \in R^{1 \times N}$, which can be obtained by adjusting $\mathcal{W}_1$. Let $\mathcal{W}_2 \in R^{|\mathcal{O}| \times 1}$ and $\bm{B} \in R^{|\mathcal{O}|}$ be undetermined parameters, which satisfies that
\begin{align}{\label{eq_b2}}
  & \tilde{\mathcal{O}}^{[j]} = \mathcal{W}_2\mathcal{O}_b^{[j]} + \bm{B}, \\
  & s.t. \ \ \varLambda(O^{[j]}) = \varLambda(\tilde{\mathcal{O}}^{[j]}). \nonumber
\end{align}
Eq. \ref{eq_b2} shows that when $\mathcal{O}_b^{[j]}$ is given, the element $\mathcal{W}_1^j$ can be used to scale it and the element $\bm{B}^j$ can be used to shift it. Assume that $\mathcal{W}_2$ is initialized as a vector whose elements are all $1$s, and $\bm B$ is initialized as a vector whose elements are all $\bm 0$s. Since both $\mathcal{W}_2$ and $\bm B$ are learnable parameters, we can easily construct that
\begin{align}{\label{eq_b3}}
  & \tilde{\mathcal{O}}^{[1]} = \mathcal{W}_2^j\mathcal{O}_b^{[1]} + \bm{B}^1,  \\
  & s.t. \ \ \tilde{\mathcal{O}}^{[1]} > 1. \nonumber
\end{align}
There are two unknown parameters and one constraint in Eq. \ref{eq_b3}, which are easily satisfied during optimization. Further, we can continue to construct the case of $j=2, \cdots, C$. Specifically, we have
\begin{align}{\label{eq_b4}}
  & \tilde{\mathcal{O}}^{[j]} = \mathcal{W}_2^j\mathcal{O}_b^{[j]} + \bm{B}^j,  \\
  & s.t. \ \ \tilde{\mathcal{O}}^{[j]} > max(\{ \tilde{\mathcal{O}}^{[k \neq j]} \}_{k=1}^C). \nonumber
\end{align}

It can be found that Eq. \ref{eq_b4} and Eq. \ref{eq_b1_1} are equivalent to each other. Thus, the proof of Theorem \ref{th3} is completed.
\end{proof}

\section{Proof of Theorem \ref{th4}}
\label{sec_a_th4}
%\vspace{-4em}

\begin{proof}
  The approximation error $E$ of the models in Eq. \ref{eq10_} can be estimated as
  \begin{align}{\label{eq_c1}}
    E & = \frac{1}{N} (\sigma(\mathcal{O}) - \sigma(\mathcal{O}^*))  \nonumber
        = \frac{1}{N} ( \sigma(\tilde{\mathcal{W}}\mathcal{F}) - \sigma(\mathcal{W}^*\mathcal{F})) \\ \nonumber
      & = \frac{1}{N} \sum_i^N \sum_j^{C}| \frac{e^{\tilde{\mathcal{W}}_j F_i}}{\sum_k{e^{\tilde{\mathcal{W}}_k \mathcal{F}_i}}}
          - \frac{e^{\mathcal{W}_j^* \mathcal{F}_i}}{\sum_k{e^{\mathcal{W}_k^* \mathcal{F}_i}}} | \\ \nonumber
      & = \frac{1}{N} \sum_i^N \sum_j^{C}| \frac{ \sum_k{e^{(\mathcal{W}_k^* + \tilde{\mathcal{W}}_j) \mathcal{F}_i}
        - \sum_k{e^{(\tilde{\mathcal{W}}_k + \mathcal{W}_j^*) \mathcal{F}_i}}}}
        {\sum_k{e^{\tilde{\mathcal{W}}_k \mathcal{F}_i}} \sum_k{e^{\mathcal{W}_k^* \mathcal{F}_i}}} | \\ \nonumber
      & = \frac{1}{N} \sum_i^N \sum_j^{C}| \frac{ \sum_k{e^{(\mathcal{W}_k^* + \mathcal{W}_j^* + \Delta_j) \mathcal{F}_i}
        - \sum_k{e^{(\mathcal{W}_k^* + \Delta_k + \mathcal{W}_j^*) \mathcal{F}_i}}}}
        {\sum_k{e^{(\mathcal{W}_k^* + \Delta_k) \mathcal{F}_i}} \sum_k{e^{\mathcal{W}_k^* \mathcal{F}_i}}} | \\ \nonumber
      & = \frac{1}{N} \sum_i^N \sum_j^{C}| \frac{ \sum_k{e^{(\mathcal{W}_k^* + \mathcal{W}_j^*) \mathcal{F}_i}} (e^{\Delta_j}
        - \sum_k{e^{\Delta_k})}}
        {\sum_k{e^{(\mathcal{W}_k^* + \Delta_k) \mathcal{F}_i}} \sum_k{e^{\mathcal{W}_k^* \mathcal{F}_i}}} | \\ \nonumber
      & = \frac{1}{N} \sum_i^N \sum_j^{C}| \frac{  e^{\mathcal{W}_j^* \mathcal{F}_i} (e^{\Delta_j}
        - \sum_k{e^{\Delta_k}})}
        {\sum_k{e^{(\mathcal{W}_k^* + \Delta_k) \mathcal{F}_i}}} | \\ \nonumber
      & \le \frac{1}{N} \sum_i^N \sum_j^{C}| e^{\Delta_j} - \sum_k{e^{\Delta_k} } | \\
      & \le \sum_j^{C} | e^{\Delta_j} - 1 |.
  \end{align}
\end{proof}

\section{}
\label{sec_af}

\begin{algorithm}[H]
  \caption{Label Decomposition}
  \label{alg_lp}
  \begin{algorithmic}[1] % line number
  \Require $\bm{Y}, \{|\bm{O}|\}_{h=1}^H$
  \State $\bm{Y}_{local} = set()$ % [\ \ ]
  \For{$h = 1 \to H$}
    \State $\bm{Y}_{sum} \gets \prod_{j=h+1}^H |\bm{O}^j|$
    \State $\bm{Y}_{cur} \gets \bm{Y} \ {\bf mod} \ \bm{Y}_{sum}$
    \State $\bm{Y} \gets \bm{Y} - \bm{Y}_{sum} \times \bm{Y}_{cur}$
    \State $\bm{Y}_{local} \gets add(\bm{Y}_{local},\bm{Y}_{cur})$
  \EndFor
  \State $\bm{Y}_{local} \gets add(\bm{Y}_{local},\bm{Y})$
  \State{\Return $\bm{Y}_{local}$}
  \end{algorithmic}
\end{algorithm}

\subsection{The Algorithm of MHP}
\label{sec_af1}
%The label partitioning algorithm is given in Algorithm \ref{alg_lp}, and the MHP algorithm is given in Algorithm \ref{alg_mhp}.

\begin{algorithm}[H]
  \caption{Multi-Head Product (MHP)}
  \label{alg_mhp}
  \begin{algorithmic}[1] % line number
  \Require $Net, \mathcal{X}, \bm{Y}, H, \{|\mathcal{O}|\}_{h=1}^H, \{\mathcal{C}^h\}_{h=1}^H, Epochs$

  \Function {$Loss$}{$\mathcal{O}_{cat}, \{\bm{Y}^h\}_{h=1}^H, H$}
        \State $\bm{Y}_{cat} = set() $
        \For{$h = 1 \to H $}
            \State $ \bm{Y}_{cat} \gets add(\bm{Y}_{cat},OHE(Y^h)) $
        \EndFor
        \State $loss \gets BCE(sigmoid(\mathcal{O}_{cat}),\bm{Y}_{cat})$
        \State \Return{$loss$}
  \EndFunction

  \State $\mathcal{C}_{cat} = Concatenate(\{\mathcal{C}^h\}_{h=1}^H)$

  \State ${\bf Training:}$
  \State $ \{\bm{Y}^h\}_{h=1}^H \gets LabelDecomposition(\bm{Y}_{train}, \{|\mathcal{O}|\}_{h=1}^H)$
  \For{$epoch = 1 \to Epochs $}
    \State $ \mathcal{F} \gets Net(\mathcal{X}_{train})$
    \State $ \mathcal{O}_{cat} \gets C_{cat} (\mathcal{F})$
    \State $ loss \gets Loss(\mathcal{O}_{cat}, \{\bm{Y}^h\}_{h=1}^H, H)$
    \State $ backward(loss)$
  \EndFor

  \State ${\bf Testing:}$
  \State $ \{\bm{Y}^h\}_{h=1}^H \gets LabelDecomposition(\bm{Y}_{test},\{|\mathcal{O}|\}_{h=1}^H)$
  \State $\tilde{\bm{Y}}_{bool} = set()$
  \For{$epoch = 1 \to Epochs $}
    \State $ \mathcal{F} \gets Net(\mathcal{X}_{test})$
    \State $ \{\mathcal{O}^h\}_{h=1}^H \gets \mathcal{C}_{cat}(\mathcal{F})$
    \State $\bm{Y}_{bool} = Vector(elements=True, length=Batch)$
    \For{$h = 1 \to H $}
      \State $\bm{Y}_{pred} \gets \varLambda(\mathcal{O}^h)$
      \State $\bm{Y}_{bool} \gets \bm{Y}_{bool} \ {\bf and} \  (\bm{Y}_{pred} \ {\bf and} \ \bm{Y}^h)$
      \State $\tilde{\bm{Y}}_{bool} \gets add(\tilde{\bm{Y}}_{bool}, \bm{Y}_{bool})$
    \EndFor
  \EndFor
  \State $Accuracy \gets \frac{ sum(\tilde{\bm{Y}}_{bool})}{|\tilde{\bm{Y}}_{bool}|}$
  \State{\Return $Accuracy$}

  \State ${\bf Predicting:}$
  \State $ \mathcal{F} \gets Net(\mathcal{X}_{test})$
  \State $ \{\mathcal{O}^h\}_{h=1}^H \gets \mathcal{C}_{cat}(\mathcal{F})$
  \State $\tilde{\bm{Y}}_{pred} = 0$
  \For{$h = 1 \to H-1 $}
    \State $\tilde{\bm{Y}}_{pred} \gets \tilde{\bm{Y}}_{pred} + \varLambda(\mathcal{O}^h) \prod_{j=h+1}^H |\mathcal{O}^j| $
  \EndFor
  \State $\tilde{\bm{Y}}_{pred} \gets \tilde{\bm{Y}}_{pred} + \varLambda(\mathcal{O}^H)  $
  \State{\Return $\tilde{\bm{Y}}_{pred}$}
  \end{algorithmic}
\end{algorithm}

\subsection{The Algorithm of MHC}
\label{sec_af2}
%The algorithm for MHE is given in Algorithm \ref{alg_mhe}.
\begin{algorithm}[H]
  \caption{Multi-Head Cascade (MHC)}
  \label{alg_mhe}
  \begin{algorithmic}[1] % line number
  \Require $Net, \mathcal{X}, \bm{Y}, H, \mathcal{E}, K, \{|\mathcal{O}|\}_{h=1}^H, \{\mathcal{C}^h\}_{h=1}^H, Epochs$
  \Function {$GetOutputs$}{$h, \mathcal{O}, \bm{Y}_{pre}, \bm{Y}^h, K$}
    \If {$\bm{Y}^h \ is \ not \ None$}
      \State $\mathcal{O} \gets \mathcal{O} + \mathbb{I}_{\bm{Y}^h} $
    \EndIf

    \State $\bm{I}_{topK} \gets Top\text{-}K(\mathcal{O},K)$
    \State $i_h \gets \prod_{j=1}^h{|\mathcal{O}^j|}$
    \State $\mathbb{C} \gets Matrix(elements=(1,...,i_h*|\mathcal{O}^{h+1}|),$
     $ shape=(i_h,|\mathcal{O}^{h+1}|))$

    \If {$\bm{Y}_{pre} \ is \ not \ None$}
      \State $\bm{I}_{topK} \gets \bm{Y}_{pre}[\mathcal{I}_{topK}]$
    \EndIf

    \State $\bm{Y}_{topK} \gets \mathbb{C}[\bm{I}_{topK}]$
    %\State $S_{topK} \gets O[I_{topK}]$
    \State \Return{$\bm{Y}_{topK}$}
  \EndFunction
  \\
  \If {$\bm{Y} \ is \ multi\text{-}hot$}
    \State $Loss = BCE$
  \Else
    \State $Loss = CE$
  \EndIf
  \State $\bm{Y}_{pre} \gets Matrix(elements=(1,...,|\mathcal{O}^1|*|\mathcal{O}^2|),shape=(|\mathcal{O}^1|,|\mathcal{O}^2|))$
  \\
  \State ${\bf Training:}$
  \State $ \{\bm{Y}^h\}_{h=1}^H \gets LabelDecomposition(\bm{Y}_{train},\{|\mathcal{O}|\}_{h=1}^H)$
  \For{$epoch = 1 \to Epochs $}
    \State $loss = 0$
    \State $ \mathcal{F} \gets Net(\mathcal{X}_{train})$
    \State $ \mathcal{O} \gets \mathcal{C}^1(\mathcal{F})$
    \State $loss \gets loss + Loss(\mathcal{O},\bm{Y}^1)$
    \For{$ h = 2 \to H $}
      \State $ \bm{Y}_{pre} \gets GetOutputs(h, \mathcal{O}, \bm{Y}_{pre}, \bm{Y}^h, K)$
      \State $\mathcal{O} \gets \mathcal{W}^h \mathcal{F} {\mathcal{E}^h}^T(\bm{Y}_{pre})$
      \State $loss \gets loss + Loss(\mathcal{O},\bm{Y}^h) $
    \EndFor
  \EndFor
  \\
  \State ${\bf Testing:}$
  \State $ \mathcal{F} \gets Net(\mathcal{X}_{test})$
  \State $ \mathcal{O} \gets \mathcal{C}^1(\mathcal{F})$
  \For{$ h = 2 \to H $}
    \State $ \bm{Y}_{pre} \gets GetOutputs(h, \mathcal{O}, \bm{Y}_{pre}, \bm{Y}^h, K)$
    \State $\mathcal{O} \gets \mathcal{W}^h\mathcal{F} {\mathcal{E}^h}^T(\bm{Y}_{pre})$
  \EndFor

  \If {$\bm{Y} \ is \ multi\text{-}hot$}
    \State $\tilde{\bm{Y}}_{pred} \gets Top\text{-}K(\mathcal{O},K)$
    \State $P@1,P@3,P@5 \gets P@K(\tilde{\bm{Y}}_{pred},\bm{Y}_{test},[1,3,5])$
    \State{\Return $\tilde{\bm{Y}}_{pred}, P@1,P@3,P@5$}
  \Else
    \State $\tilde{\bm{Y}}_{pred} \gets \varLambda(\mathcal{O})$
    \State $\bm{Y}_{bool} \gets \tilde{Y}_{pred} \ {\bf and} \ \bm{Y}_{test}$
    \State $Accuracy \gets \frac{sum(\bm{Y}_{bool})}{|\bm{Y}_{bool}|}$
    \State{\Return $\tilde{\bm{Y}}_{pred}, Accuracy$}
  \EndIf
  \end{algorithmic}
\end{algorithm}

\subsection{The Algorithm of MHS}
\label{sec_af3}

\begin{algorithm}[H]
  \caption{Multi-Head Sampling (MHS)}
  \label{alg_mhs}
  \begin{algorithmic}[1] % line number
  \Require $Net, \mathcal{X}, \bm{Y}, S, \{|\mathcal{O}|\}_{h=1}^H, Epochs$

  \Function{$LabelPartitioning$}{$\bm{Y}_{train}, \{|\mathcal{O}^h|\}_{h=1}^H$}
    \State $\bm{Y}_{head} = set()$
    \State $\bm{Y}_{local} = set()$
    \For{$h = 1 \to H $}
      \State $\bm{Y}_{head} \gets add(\bm{Y}_{head}, h)$
      \If {$h > 1$}
        \State $\bm{Y}_{local} \gets add(\bm{Y}_{local}, \bm{Y}_{train} - \sum_{i=1}^{h-1} |\mathcal{O}^i|) $
      \Else
        \State $\bm{Y}_{local} \gets add(\bm{Y}_{local}, \bm{Y}_{train}) $
      \EndIf
    \EndFor
    \State \Return{$\bm{Y}_{head}, \bm{Y}_{local}$}
  \EndFunction
  \Function {$Sampling$}{$\mathcal{F}, \bm{Y}_{local}, S, H $}
    \State $\mathcal{O} = set()$
    \For{$i = 1 \to |\mathcal{F}|$}
      \State $\mathcal{C}_{pos} \gets \bm{Y}_{local}^i$
      \State $\bm{I} \gets Vector([1,...,\mathcal{C}_{pos}-1,\mathcal{C}_{pos}+1,...,H])$
      \State $\bm{I} \gets shuffle(I)$
      \State $\bm{I}[0] \gets \mathcal{C}_{pos}$
      \State $\mathcal{C}_{select} \gets \mathcal{C}[I[:S]]$
      \State $\mathcal{O} \gets add(\mathcal{O},\mathcal{C}_{select}(\mathcal{F}[i]))$
    \EndFor
    \State \Return{$\mathcal{O}, \bm{Y}_{local}$}
  \EndFunction
  \Function{$FastSampling$}{$\mathcal{F}, \bm{Y}_{head}, \bm{Y}_{local}$}
    \State $\bm{Y}_{uni} \gets Unique(\bm{Y}_{head})$
    \State $\bm{I}_{pos} \gets  Searchsorted(\bm{Y}_{uni}, \bm{Y}_{head})$
    \State $\bm{Y}_{new} \gets  \bm{I}_{pos} * \bm{Y}_{head} +  \bm{Y}_{local}$
    \State $ \mathcal{W}_{select} \gets \mathcal{W}[\bm{Y}_{uni}]  $
    \State $ \mathcal{O} = \mathcal{W}_{select}\mathcal{F} $
    \State \Return{$\mathcal{O}, \bm{Y}_{new}$}
  \EndFunction
  \State $  \bm{Y}_{head}, \bm{Y}_{local} \gets LabelPartitioning(\bm{Y}_{train},\{|\mathcal{O}|\}_{h=1}^H)$
  \\
  \State ${\bf Training:}$
  \For{$epoch = 1 \to Epochs $}
    \State $ \mathcal{F} \gets Net(\mathcal{X}_{train})$
    \State $ \mathcal{O}, \mathcal{Y}_{new} \gets FastSampling(\mathcal{F}, \bm{Y}^h, S)$
    \State $ loss \gets \Call{Loss}{\mathcal{O}, \bm{Y}_{new}}$
    \State $ backward(loss)$
  \EndFor
  \State ${\bf Testing (if \ necessary):}$
  \State $ \mathcal{F} \gets Net(\mathcal{X}_{test})$
  \State $ \mathcal{O} \gets \mathcal{W}\mathcal{F}$
  \State $\tilde{\bm{Y}}_{pred} \gets \varLambda(\mathcal{O})$
  \State $\bm{Y}_{bool} \gets \tilde{\bm{Y}}_{pred} \ {\bf and} \ \bm{Y}_{test}$
  \State $Accuracy \gets \frac{sum(\bm{Y}_{bool})}{|\bm{Y}_{bool}|} $
  \State{\Return $\tilde{\bm{Y}}_{pred}, Accuracy$}
  \end{algorithmic}
\end{algorithm}

\clearpage

\section{Extra Experimental Results}
\label{sec_ae}

\subsection{Example of the Kronecker Product}
\label{sec_ae_kron}

\begin{figure}[h]
  \centering
  \includegraphics[width=0.4\textwidth]{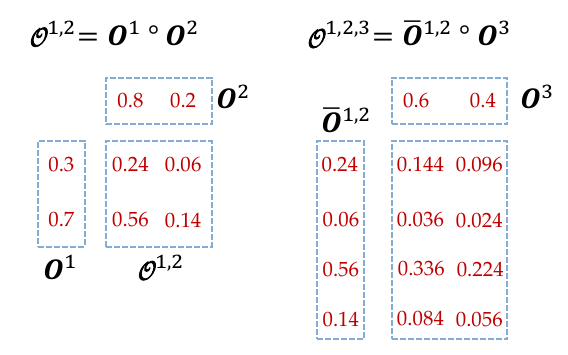}
  \caption{
    The example of the Kronecker product used in MHE. The symbol $\circ$ denotes the outer product of two vectors.
  }
  \label{fig_r1_9}
\end{figure}

For the example of the Kronecker product used in MHE, some examples are given to explain the elusive operations. 
The essence of the Kronecker product operation is to vectorize the outer product of multiple vectors. For example, assume that the outputs of the three heads are $\bm{O}^1$=[0.3,0.7], $\bm{O}^2$=[0.8,0.2] and $\bm{O}^3$=[0.6,0.4], respectively, and we want to calculate their Kronecker product result $\bar{\bm O} = \bm{O}^1 \otimes \bm{O}^2 \otimes \bm{O}^3$. The calculation process is shown in Fig. \ref{fig_r1_9}. 
It is noted that the outer product of two vectors is equivalent to their Cartesian product. Then, the outer product operation is again performed on the vectorized $\mathcal{O}^{1,2}$ and $\bm{O}^3$. 
Finally, $\mathcal{O}^{1,2,3}$ is vectorized to get the final output. 

% The reshaping operation in the mapping process is to rearrange the vector $\bar{\bm Y}_i$ into an $H$-order tensor $\mathcal{Y}_i^{1,\cdots,H}$ according to its indexes. 
% e.g., $\mathcal{Y}_i^{1,2,3}$ = [[[0,0],[0,0]],[[1,0],[0,0]]] (numpy-style). Now, we get the three coordinate components (sublabels) of element 1 in $\mathcal{Y}_i^{1,2,3}$. 
% Further, if $C=9$ and the lengths of the two heads are all set to be 3, then $\bar{\bm Y}_i$ = [0,0,0,0,1,0,0,0,0], $\mathcal{Y}_i^{1,2}$ = [[0,0,0],[0,1,0], [0,0,0]]. Finally, we have $\bar{\bm{Y}}^1$ = [0,0,1] and $\bar{\bm{Y}}^2$ = [0,1,0].

\subsection{Implementation of Multi-Head Classifier}
\label{sec_ae_hyper}

The proposed methods have several hyperparameters to be determined, including the number of classification heads and their lengths.  
In practice, the number of classifiers and their lengths can be determined according to the principles proposed in Section \ref{sec4_4}. 
Specifically, assuming that we have $H$ classification heads $\{\mathcal{C}^h\}_{h=1}^H$, their outputs $\{\bm{O}^h\}_{h=1}^H$ and weights $\{\mathcal{W}^h\}_{h=1}^H$ can be combined as
\begin{align}
  \bar{\bm{O}} & = \{\bm{O}^1, \bm{O}^2, \cdots ,\bm{O}^H \} \nonumber \\
  & = \{\mathcal{W}^1F, \mathcal{W}^2F, \cdots ,\mathcal{W}^HF \} \nonumber \\  
  & = \{\mathcal{W}^1, \mathcal{W}^2, \cdots ,\mathcal{W}^H \}F \nonumber \\
  & = \bar{\mathcal{W}}F
\end{align}
where $\bar{\bm{O}} $ and $\bar{\mathcal{W}}$ are the concatenated outputs $\{\bm{O}^h\}_{h=1}^H$ and weights  $\{\mathcal{W}^h\}_{h=1}^H$. Let $L = \sum_{h=1}^H |\mathcal{C}^h|$ denote the total length of all heads, we have $|\bar{\mathcal{W}}| = L$. 
In the implementation of the multi-head classifier, the outputs $\bar{\bm{O}}$ are obtained by matrix multiplication of weights $\bar{\mathcal{W}}$ and features $\bm F$. 
Therefore, reducing the total length $L$ is equivalent to reducing the number of parameters of the multi-head classifier, thereby reducing the computational consumption. 

Furthermore, the length of each head is determined according to the principle of confusion degree. Specifically, when $H$ is determined by the principles of error accumulation, we have $|\mathcal{C}^h| \approx \sqrt[H]{C} $, where $C$ is the number of categories. For example, if $C$ = 1728000 and $H$=3. Then, the length of each head is $\sqrt[3]{C}$=120 and the total parameters in the classifier are 360$|\bm F|$. 
While $H$=4, the length of each head is $\sqrt[4]{C} \approx $36 and the total parameters in the classifier are 144$|\bm F|$.
% Another way to determine the length of each head is to factor the categories $C$. Then, the factors are combined for the purpose of reducing the total length $L$. e.g., on Amazon3M dataset, $C$=2812281, factoring it out gives $C= 3 \times 107 \times 8761$. Therefore, we can determine that the lengths of the two heads are 321 and 8761, respectively. 
% Certainly, we can divide it into 3 heads to further reduce computing resources. e.g., $C \approx 128 \times 127 \times 173$.    

\subsection{Experiments on Commonly used Image Datasets}

Alternatively, we perform warm-up experiments on the ImageNet-2012 \cite{russakovsky2015imagenet}, CIFAR-10 \cite{Krizhevsky2009Learning}, and CIFAR-100 \cite{Krizhevsky2009Learning} datasets. The purpose of these experiments is to verify the effectiveness of MHE-based algorithms.
It can be seen from Table \ref{tb_cafir} that the performances of the three MHE-based algorithms ($H\ge 2$) is close to that of the vanilla method ($H=1$). It can also be seen from Table \ref{tb1} that the proposed MHE-based algorithms outperform the three existing SOTA methods in solving the XLC problem, including AttentionXML \cite{you2019attentionxml}, X-Transformer \cite{chang2020taming}, and LightXML \cite{jiang2021lightxml}. The experimental results further demonstrate the flexibility and applicability of MHE, paving a way for solving various XLC tasks.

\begin{table}
  \centering
  \caption{\footnotesize Experiments on commonly used image datasets.}
  \label{tb_cafir}
  \begin{threeparttable}

   \begin{subtable}{\columnwidth}
    \vspace{0.5 em}
    % \scriptsize
    \resizebox{\columnwidth}{!}
      {
        \begin{tabular}{c|cccc|ccc}
          \hline
          \multicolumn{1}{c|}{Dataset}  & \multicolumn{3}{c|}{CIFAR-10}                                                                                                   & \multicolumn{3}{c}{ImageNet}         \\ \hline
          Methods         &  Setting                               & 200                      & 400                        & \multicolumn{2}{|c}{Setting}                 & 90                                    \\ \hline
          Vanilla         &  H={\{10\}}                            & 95.57{\tiny $\pm 0.22$}  & 95.73{\tiny $\pm 0.16$}    & \multicolumn{2}{|c}{H={\{1000\}}}            & 73.16{\tiny $\pm 0.18$}          \\  \hline
          AttentionXML    &  HLT={\{2;5\}}                         & 93.34{\tiny $\pm 0.35$}  & 93.56{\tiny $\pm 0.32$}    & \multicolumn{2}{|c}{HLT={\{4;8;32\}}}        & 66.38{\tiny $\pm 0.31$}          \\ 
          X-Transformer   &  LC={\{5;2\}}                          & 93.78{\tiny $\pm 0.31$}  & 94.07{\tiny $\pm 0.28$}    & \multicolumn{2}{|c}{LC={\{40;25\}}}          & 67.54{\tiny $\pm 0.42$}          \\ 
          LightXML        &  LC={\{5;2\}}                          & 94.15{\tiny $\pm 0.26$}  & 94.43{\tiny $\pm 0.23$}    & \multicolumn{2}{|c}{LC={\{40;25\}}}          & 68.49{\tiny $\pm 0.37$}          \\  
          MHP             &  H={\{5;2\}}                           & 95.05{\tiny $\pm 0.19$}  & 95.31{\tiny $\pm 0.15$}    & \multicolumn{2}{|c}{H={\{40;25\}}}           & 70.34{\tiny $\pm 0.24$}          \\ 
          MHC             &  H={\{5;2\}}                           & 95.26{\tiny $\pm 0.14$}  & 95.55{\tiny $\pm 0.12$}    & \multicolumn{2}{|c}{H={\{40;25\}}} & 72.06{\tiny $\pm 0.21$}          \\
          MHS             &  H={\{5;2\}}                           & 94.73{\tiny $\pm 0.17$}  & 95.28{\tiny $\pm 0.10$}    & \multicolumn{2}{|c}{H={\{40;25\}}}           & \bf{73.51}{\tiny $\pm 0.18$}     \\ \hline
        \end{tabular}
      }
  \end{subtable}

  \begin{subtable}{\columnwidth}
    \vspace{0.5 em}
    \Large
    \resizebox{\columnwidth}{!}
    {
      \begin{tabular}{c|cccccc}
        \toprule
        \multicolumn{1}{c|}{Dataset}      & \multicolumn{5}{c}{CIFAR-100}                                                                                                                                                    \\ \hline
        Methods         & Setting                     & \multicolumn{1}{c|}{400}                          & Setting            & \multicolumn{1}{c|}{400}                             & Setting                    & 400   \\ \hline
        Vanilla         & H={\{100\}}                 & \multicolumn{1}{c|}{77.70{\small $\pm 0.13$}}     & H={\{100\}}        & \multicolumn{1}{c|}{77.70{\small $\pm 0.13$}}        & H={\{100\}}                & 77.70{\small $\pm 0.13$}  \\  \hline
        AttentionXML    & HLT={\{4;25\}}              & \multicolumn{1}{c|}{73.24{\small $\pm 0.22$}}     & HLT={\{8;13\}}     & \multicolumn{1}{c|}{71.57{\small $\pm 0.34$}}        & HLT={\{4;4;7\}}            & 72.14{\small $\pm 0.31$}  \\ 
        X-Transformer   & LC={\{20;5\}}               & \multicolumn{1}{c|}{73.96{\small $\pm 0.24$}}     & LC={\{10;10\}}     & \multicolumn{1}{c|}{72.45{\small $\pm 0.26$}}        & LC={\{5;20\}}              & 73.17{\small $\pm 0.26$}  \\ 
        LightXML        & LC={\{20;5\}}               & \multicolumn{1}{c|}{74.07{\small $\pm 0.17$}}     & LC={\{10;10\}}     & \multicolumn{1}{c|}{69.90{\small $\pm 0.27$}}        & LC={\{5;20\}}              & 73.66{\small $\pm 0.28$}  \\  
        MHP             & H={\{20;5\}}               & \multicolumn{1}{c|}{74.84{\small $\pm 0.14$}}      & H={\{10;10\}}      & \multicolumn{1}{c|}{75.92{\small $\pm 0.15$}}        & H={\{4;5;5\}}              & 74.53{\small $\pm 0.16$}  \\
        MHC             & H={\{20;5\}}               & \multicolumn{1}{c|}{77.36{\small $\pm 0.15$}}      & H={\{10;10\}}      & \multicolumn{1}{c|}{76.26{\small $\pm 0.13$}}        & H={\{4;5;5\}}              & 76.11{\small $\pm 0.13$}  \\
        MHS             & H={\{20;5\}}              & \multicolumn{1}{c|}{\bf{78.51}{\small $\pm 0.14$}}  & H={\{10;10\}}      & \multicolumn{1}{c|}{\bf{77.95}{\small $\pm 0.14$}}   & H={\{4;25\}}               & \bf{77.83}{\small $\pm 0.16$} \\  \bottomrule
      \end{tabular}
     }
  \end{subtable}

  \scriptsize
  \begin{tablenotes}
      \item[*] The numbers in the set $\{\cdot \ ;\  \cdot \}$ indicate the setting of different XLC methods. \\`H' and `LC' represent the settings of heads and label clustering respectively.
  \end{tablenotes}
  \end{threeparttable}
\end{table}

\subsection{Extra Experiments of Label Decomposition}
\label{subsec_a_lrd}

\begin{figure*}
  \centering
  \begin{minipage}{\textwidth}
    \includegraphics[width=\textwidth]{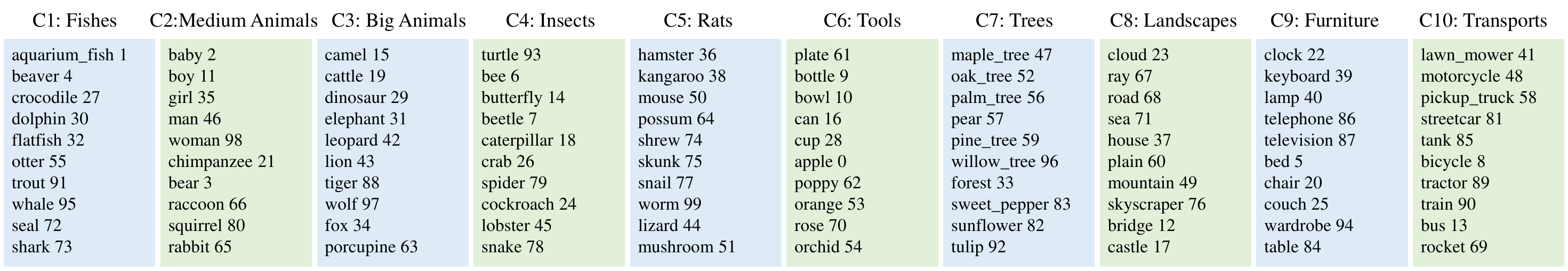}
    % \captionsetup{font={scriptsize},justification=justified}
    \caption{
      The labels in CIFAR-100 are clustered into 10 clusters, each of which contains 10 categories.
    }
  \label{fig_r1q1_2}
  \end{minipage}
\end{figure*}

\begin{table*}[!ht]
  % \captionsetup{font={scriptsize},justification=justified}
  \caption{Experiments of label rearrangement and decomposition equivalence. LRD indicates label random rearrangement and decomposition, and LC indicates preprocessing with label clustering.}
  \centering
  \label{tb_lrd}
  \large
  \resizebox{\textwidth}{!}
  {
    \begin{tabular}{c|cccc|cccccccc|cc}
      \toprule                            
            & \multicolumn{4}{c|}{CIFAR-10}                                                                         & \multicolumn{8}{c|}{CIFAR-100}                                                                                                                                                                                                                                                                                 & \multicolumn{2}{c}{Eurlex-4K}     \\ \midrule
      Head  & \multicolumn{2}{c|}{\{5;2\}+LC}                                                  & \multicolumn{2}{c|}{\{5;2\}+LRD}   & \multicolumn{2}{c|}{\{20;5\}+LC}        & \multicolumn{2}{c|}{\{20;5\}+LRD}          & \multicolumn{2}{c|}{\{10;10\}+LC}                                                                                           & \multicolumn{2}{c|}{\{10;10\}+LRD}      &  \multicolumn{1}{c|}{\{172;23\}+LC}                                                              & {\{172;23\}+LRD}                    \\ \midrule
      Epoch & 200                        & \multicolumn{1}{c|}{400}                         & 200                       & 400                       & 200                                       & \multicolumn{1}{c|}{400}                      & 200                       & \multicolumn{1}{c|}{400}                      & 200                     & \multicolumn{1}{c|}{400}                        & 200             & 400       & \multicolumn{1}{c|}{25}          & 25          \\ \midrule
      MHP   & 95.05{\small $\pm 0.19$}   & \multicolumn{1}{c|}{95.31{\small $\pm 0.15$}}    & 95.35{\small $\pm 0.17$}   & 95.55{\small $\pm 0.14$}   & \multirow{1}{*}{74.30{\small $\pm 0.18$}}  & \multicolumn{1}{c|}{74.84{\small $\pm 0.14$}}  & 74.24{\small $\pm 0.12$}   & \multicolumn{1}{c|}{74.29{\small $\pm 0.21$}}  & 75.47{\small $\pm 0.15$} & \multicolumn{1}{c|}{75.92{\small $\pm 0.15$}}    & 75.48{\small $\pm 0.18$}     & 75.96{\small $\pm 0.16$}           & \multicolumn{1}{c|}{86.00{\small $\pm 0.16$} }       & 85.95{\small $\pm 0.17$}     \\
      MHE   & 95.26{\small $\pm 0.14$}   & \multicolumn{1}{c|}{95.55{\small $\pm 0.12$}}    & 95.41{\small $\pm 0.17$}   & 95.62{\small $\pm 0.12$}   & \multirow{1}{*}{76.73{\small $\pm 0.16$}}  & \multicolumn{1}{c|}{77.36{\small $\pm 0.15$}}  & 76.73{\small $\pm 0.16$}   & \multicolumn{1}{c|}{77.45{\small $\pm 0.15$}}  & 75.93{\small $\pm 0.14$} & \multicolumn{1}{c|}{76.26{\small $\pm 0.13$}}    & 76.03{\small $\pm 0.14$}     & 76.25{\small $\pm 0.17$}           & \multicolumn{1}{c|}{74.39{\small $\pm 0.13$} }       & 74.31{\small $\pm 0.14$}      \\
      MHS   & 94.73{\small $\pm 0.17$}   & \multicolumn{1}{c|}{95.28{\small $\pm 0.10$}}    & 94.77{\small $\pm 0.15$}   & 95.24{\small $\pm 0.13$}   & \multirow{1}{*}{77.46{\small $\pm 0.13$}}  & \multicolumn{1}{c|}{78.51{\small $\pm 0.14$}}  & 77.49{\small $\pm 0.15$}   & \multicolumn{1}{c|}{78.56{\small $\pm 0.13$}}  & 77.05{\small $\pm 0.12$} & \multicolumn{1}{c|}{77.95{\small $\pm 0.14$}}    & 77.10{\small $\pm 0.18$}     & 77.91{\small $\pm 0.16$}          & \multicolumn{1}{c|}{62.05{\small $\pm 0.18$} }       & 62.02{\small $\pm 0.15$}      \\ \bottomrule
    \end{tabular}
  }

\end{table*}

In Table \ref{tb_lrd}, we compare the performance of the methods with label clustering (LC) and label rearrangement and decomposition (LRD) on the CIFAR-10, CIFAR-100, and Eurlex-4K datasets.
It should be noted that LRD involves first randomly arranging labels and then dividing them according to the number and length of heads (clusters).
For preprocessing techniques, we carefully select semantically similar labels for the same cluster, e.g., labels in CIFAR-10 are clustered into \{plane, car, bird, ship, truck\} (birds are similar to planes) and \{cat, deer, dog, frog, horse\}, and label clustering in CIFAR-100 is shown in Fig. \ref{fig_r1q1_2}. 
As shown in Table \ref{tb_lrd}, the performance of the model with label preprocessing by LRD is essentially identical to that of the model with label clustering.

The above conclusion also holds when the classifiers share parameters in the same stage. However, in this case, misclassification errors between different label partitions may occur. For instance, the second-stage classifier in Fig. \ref{fig_r1q1}.a might classify the ant as a dog.
To measure this error, we introduce the metric of confusion degree in Section \ref{sec4_4}.
It is worth noting that confusion degree is solely related to the number and length of classification heads (or clusters) and is irrelevant to the specific label arrangement.
Therefore, when the number of clusters and the number of samples they contain are fixed, the generalization of the classifiers is roughly the same.

\subsection{Long-Tailed Label Distribution in XMLC}
\label{sec_ae_longlbl}

The long-tailed label distribution is an important issue in XMLC.
To verify the influence of the long-tail distribution on the proposed method, experiments on Eurlex-4K and Wiki10-31k datasets using the proposed MHC method are conducted by trimming tail labels. The experimental results in Table \ref{ap_tb1} show that when labels share equal weights, the impact of the tail labels is much less than that of the common labels, which is consistent with
the conclusions in the literature \cite{tail_label_8830456}. 

Specifically, when 25\% of the tail labels are trimmed from the training set of Eurlex-4K, the performance of MHC only drops by 0.2\%. However, when 50\% of the tail labels are trimmed, the performance of MHC drops by 50\%. This suggests that trimming less performance-influential labels has little impact on the final performance of the model. For example, the binary search algorithm is developed in \cite{tail_label_8830456} to efficiently determine the cutoff threshold based on observation performance.

On the Wiki10-31k dataset, it is found in Table \ref{ap_tb1} that the performance of the model is only slightly affected even when 75\% of the tail labels are trimmed off. Interestingly, the P@1 score after 75\% trimming is higher than after 50\% trimming, while the opposite is true for the P@3 and P@5 scores. 
In summary, there is still a lot of interesting and worthwhile work to be explored regarding the long-tail distribution problem in extreme multi-label learning. For example, there may be further performance improvements if some balanced loss functions are developed for the long-tail distribution problem. We will also follow up on this research further.

\begin{table}[!h]
  \centering 
  \caption{Performance of MHC after trimming tail labels with different ratios on Eurlex-4K and Wiki10-31k datasets.}
  \label{ap_tb1}
  \resizebox{\columnwidth}{!}
  { \Large
    \begin{tabular}{c|ccc|ccc}
    \toprule
    Dataset        & \multicolumn{3}{c|}{Eurlex-4K, MHC\{172;23\}}       & \multicolumn{3}{c}{Wiki10-31k, MHC\{499,62\}}  \\ \midrule
    \multirow{1}{*}{Trimming Rate}   & 0\%      & 25\%     & 50\%       & 0\%       & 50\%     & 75\%  \\ \midrule
                      P@1      & 86.00     & 85.80    & 43.13      & 89.40     & 88.60    & 88.72  \\
                      P@3      & 74.39     & 72.64    & 28.82      & 79.22     & 77.70    & 77.54  \\
                      P@5      & 62.05     & 60.32    & 21.05      & 70.25     & 68.16    & 67.32  \\ \bottomrule
    \end{tabular}
  }
\end{table}

\section{Extra Experimental Setting}
\label{app_ag}

\subsection{MHE for XSLC}
\label{sec_ae1}

For the CIFAR datasets, standard data augmentation methods such as horizontal flipping and translation by 4 pixels are adopted. ResNet-18 is used as the backbone model. The learning rate of the optimizer SGD is set to be 0.1 and gradually reduced to 1e-4 using the cosine annealing scheduler.

The ImageNet dataset contains 1.2 million images for training, and 50K for verification. We use the same data augmentation scheme as in the work \cite{He2016ResNet} for the training images and apply 224$\times$224 center cropping to the images during testing. ResNet-50 \footnote[1]{The model is implemented by www.pytorch.org} is used as the backbone model and is trained on 8 Tesla V100 GPUs. The learning rate of the optimizer SGD is set to 0.1 and drops to 10\% every 30 epochs.

\subsection{MHC for XMLC}
\label{sec_ae2}

For XMLC benchmarking datasets, as shown in Table \ref{ap_tb2}, we directly use raw text without any preprocessing. The dropout in the classifier is set to be 0.5, and the weight decay is set to 0.01 for the bias and weights of layer normalization. For datasets with small labels, e.g., Eurlex-4k, Amazoncat-13k, and Wiki10-31k, the case of $H=1$ is used during model ensemble, as done in many works \cite{you2019attentionxml,jiang2021lightxml}. P@K is utilized as the evaluation metric, which is widely used in XMLC tasks to represent the percentage of accurate labels in the Top-$K$ predicted labels. The batch size is set to 8, 16, or 32, which are adjusted according to the memory of the GPU and the length of the input token. Most of the models can be trained on a single Tesla V100 GPU. However, to speed up the training process, 8 Tesla V100 GPUs are also used for ensemble learning.

\begin{table}[]
  \centering
  \caption{The statistics information of the six XMLC benchmarking datasets. $N_{train}$ and $N_{test}$ denote the number of instances in the training and test sets, respectively, $N_{label}$ denotes the number of labels, $\bar{N}_{label}$ denotes the average number of positive labels per instance, $\bar{N}_{sample}$ denotes the average number of instances per label, $N_{token}$ denotes the length of tokens used in model training and testing.}
  \label{ap_tb2}
  \resizebox{\columnwidth}{!}
  { \huge
    \begin{tabular}{ccccccc}
      \toprule
       Dataset   & $N_{train}$ & $N_{test}$ & $N_{label}$ & $\bar{N}_{label}$ & $\bar{N}_{sample}$ & $N_{token}$ \\
      \midrule
      Eurlex-4K  &  15,449 & 3,865 & 3,956 & 5.3 & 20.79 & 512 \\
      Wiki10-31K  & 14,146 & 6,616 & 30,938 & 18.64 & 8.52 & 512 \\
      AmazonCat-13K  & 1,186,239 & 306,782  & 13,330 & 5.04 & 448.57 & 512\\
      Amazon-670K   & 490,449 & 153,025  & 670,091 & 5.45 & 3.99 & 128\\
      Wiki-500K   & 1,779,881 & 769,421  & 501,070 & 4.75 & 16.86 & 128\\
      Amazon-3M  & 1,717,899 & 742,507  & 2,812,281 & 36.04 & 22.02 & 128 \\
      \bottomrule
    \end{tabular}
  }
\end{table}

\subsection{MHS for Model Pretraining}
\label{sec_ae3}

For the CASIA dataset, ResNet-18 is adopted as the backbone. The lengths of the classification heads are set to 12 and 881, respectively, and the dimension of the embedding feature is set to 512. The learning rate of the optimizer SGD is set to 0.2 and gradually decreased using the Poly scheduler. We use a weight decay of 5e-4 and Nesterov momentum of 0.9 without dampening. The batch size on each GPU is set to 128 for 25 epochs.

For the MS1MV2 and MS1MV3 datasets, ResNet-101 is adopted as the backbone. The lengths of the classification heads are set to 43 and 1994, 13 and 7187 for MS1MV2 and MS1MV3 datasets, respectively. The feature scale $s$ is set to 64 and the arccos margin $m$ of ArcFace is set to 0.5. The other experimental setups are same as those for the CASIA dataset.

\subsection{MHC and MHS for NMT}
\label{sec_ae4}

For the WMT16 dataset, the OPUS-MT model is adopted for the ro-en and de-en translation tasks. It is trained on 8 Tesla V100 GPUs using AdamW. The batch size on each GPU is set to 4 for 3 epochs. During testing, the batch size on each GPU is increased to 8, and the greedy search strategy is used for prediction. The initial learning rate is set to 5e-5 and is reduced linearly over the total number of training epochs. The maximum input sequence length of the source text is set to 512, and the maximum sequence length of the target text is set to 128.

\subsection{Potential Challenges and Solutions}
\label{sec_ag5}

 MHC's cascade design could create dependencies between heads, potentially complicating both the training process and scalability in large-scale applications. To address this, we suggest several strategies to mitigate these challenges:

1) Head Reduction: The number of classification heads should be minimized when computing resources and running speed permit, as analyzed in Section \ref{sec4_4} (Label Decomposition Principle). Therefore, reducing the number of heads can avoid dependencies between heads and improve  the computational efficiency.
 
2) Parallel Training: Where feasible, partial parallelization of training between cascade layers can help reduce dependency issues. For instance, each head is assigned to an individual GPU. By sharing certain feature representations across heads or limiting information flow between layers to essential parameters, the cascade operation can be made more efficient.
 
3) Alternative Optimization Techniques: To improve training efficiency, techniques such as asynchronous optimization and gradient averaging across heads may also be exploited, as these can limit inter-head dependency effects during training.

The solutions mentioned above will better contextualize the applicability of our algorithms in real-world settings.

\end{document}